\documentclass[11pt]{article}

\usepackage[margin=1in]{geometry}
\usepackage[T1]{fontenc}
\usepackage[utf8]{inputenc}
\usepackage{lmodern}
\usepackage{microtype}
\usepackage{parskip}

\usepackage{amsmath,amssymb,amsthm,mathtools,bm}
\usepackage{booktabs,tabularx,multirow,array}
\usepackage{graphicx}
\usepackage{xcolor}
\usepackage{enumitem}
\usepackage{algorithm}
\usepackage{algorithmic}
\usepackage{upquote}
\usepackage{listings}
\usepackage{caption}
\usepackage{subcaption}
\usepackage{tikz}
\usepackage{bbm}
\usetikzlibrary{arrows.meta,positioning,calc,shapes.geometric}

\usepackage[round,authoryear]{natbib}
\usepackage{titletoc}
\usepackage[colorlinks=true,linkcolor=blue!50!black,citecolor=blue!50!black,urlcolor=blue!50!black]{hyperref}
\usepackage[capitalise,noabbrev]{cleveref}

\newcommand{\pid}{\texorpdfstring{\textcolor{red!75!black}{\texttt{partialid}}}{partialid}}

\newcommand{\pdeg}{\mathrm{deg}}

\newcommand{\ind}{\mathbf{1}}

\DeclareMathOperator{\An}{An}
\DeclareMathOperator{\Pa}{Pa}

\Crefname{equation}{Eq.}{Eqs.}
\Crefname{figure}{Fig.}{Figs.}
\Crefname{table}{Tab.}{Tabs.}
\Crefname{theorem}{Thm.}{Thms.}
\Crefname{lemma}{Lem.}{Lems.}
\Crefname{proposition}{Prop.}{Props.}
\Crefname{definition}{Def.}{Defs.}
\Crefname{algorithm}{Alg.}{Algs.}
\Crefname{corollary}{Cor.}{Cors.}
\Crefname{section}{Sec.}{Secs.}
\Crefname{condition}{Cond.}{Conds.}

\newtheorem{lemma}{Lemma}
\newtheorem{theorem}{Theorem}
\newtheorem{corollary}{Corollary}
\newtheorem{proposition}{Proposition}

\theoremstyle{definition}
\newtheorem{definition}{Definition}
\newtheorem{example}{Example}

\theoremstyle{remark}

\newcommand{\I}{\mathbbm{1}}

\newcommand{\G}{\mathcal{G}}
\newcommand{\M}{\mathcal{M}}

\def\*#1{\mathbf{#1}}
\def\1#1{\mathcal{#1}}
\def\2#1{\mathscr{#1}}
\def\3#1{\mathbb{#1}}

\DeclareBoldMathCommand{\u}{u}
\DeclareBoldMathCommand{\U}{U}
\DeclareBoldMathCommand{\v}{v}
\DeclareBoldMathCommand{\V}{V}
\DeclareBoldMathCommand{\x}{x}
\DeclareBoldMathCommand{\X}{X}
\DeclareBoldMathCommand{\y}{y}
\DeclareBoldMathCommand{\Y}{Y}
\DeclareBoldMathCommand{\z}{z}
\DeclareBoldMathCommand{\Z}{Z}
\DeclareBoldMathCommand{\c}{c}
\DeclareBoldMathCommand{\C}{C}
\DeclareBoldMathCommand{\r}{r}
\DeclareBoldMathCommand{\R}{R}
\DeclareBoldMathCommand{\s}{s}
\DeclareBoldMathCommand{\S}{S}
\DeclareBoldMathCommand{\w}{w}
\DeclareBoldMathCommand{\W}{W}
\DeclareBoldMathCommand{\t}{t}
\DeclareBoldMathCommand{\T}{T}
\DeclareBoldMathCommand{\A}{A}
\DeclareBoldMathCommand{\a}{a}
\DeclareBoldMathCommand{\B}{B}
\DeclareBoldMathCommand{\b}{b}
\DeclareBoldMathCommand{\d}{d}
\DeclareBoldMathCommand{\q}{q}

\newcommand{\supp}{\mathrm{supp}}

\lstdefinestyle{py}{
  language=Python,
  basicstyle=\ttfamily\small,
  keywordstyle=\bfseries\color{blue!45!black},
  commentstyle=\itshape\color{black!45},
  stringstyle=\color{black!70},
  showstringspaces=false,
  columns=flexible,
  keepspaces=true,
  breaklines=true,
  breakindent=1.2em,
  upquote=true,
  frame=leftline,
  framerule=1.6pt,
  framesep=0.7em,
  rulecolor=\color{blue!45!black},
  xleftmargin=1.5em,
  xrightmargin=0.2em,
  aboveskip=0.75em,
  belowskip=0.55em,
  morekeywords={True,False,None},
  literate={partialid}{{\textcolor{red!75!black}{partialid}}}{9},
}
\title{A Graphical Criterion for Linear Causal Bounds}

\author{Alexis Bellot}

\date{}

\begin{document}
\maketitle

\begin{abstract}
The values of causal effects and counterfactuals are typically only partially identified from observational data in the presence of unobserved confounding. With discretely-valued variables, under certain conditions, sharp bounds can be derived by solving one linear program. This is then computationally tractable for many causal questions. In this paper, for every diagram in which each confounded component has at most two hidden causes, we give necessary and sufficient conditions for the sharp bounds to be equal to the optimum of one linear program for every (positive) input probability table. This algorithm is implemented in a new Python package, called \pid{}, that allows users to compose assumptions, queries, and data to obtain sharp bounds in a computationally tractable manner.
\end{abstract}


\section{Introduction}
\label{sec:intro}
In the background of any causal inference problem is an impossibility result:
roughly, the idea that one may not determine the effect of interventions and counterfactuals from observational information alone \citep{pearl2009causality,bareinboim2022pearl}. It is well known that some qualitative understanding of the phenomenon is necessary for deriving the value of causal queries from data.

That understanding can take the form of functional dependencies between variables, e.g., encoded in a causal diagram, as topological orderings between variables, restrictions on the shape of functions or on the strength of unobserved confounding, among others. Important results in the causal inference community show how one might exploit this knowledge to express causal queries in terms of observational probabilities \citep{shpitser2006identification,pearl2009causality}, and narrow the bounds that are implied by the data \citep{balke1994bounds,balke1997bounds,zhang2022partial,duarte2024automated,bellot2023towards,bellot2024towards,symbolicbounds2023general,shridharan2023scalable}. %And could be combined with assumptions placed on the shape of the underlying causal mechanisms themselves, for example stating monotonicity of the treatment response to encode the fact that the treatment is always beneficial \citep{manski1997monotone}, or with various sensitivity paramters to set a maximum strength on the influence of unobserved confounders \citep{rosenbaum2002observational,tan2006distributional,masten2018identification}. %Different implementations of these ideas have lead to a large number of frameworks to study the causal inference problem and its sensitivity to assumptions; that can often be difficult to navigate and compare for a practitioner.

When the data is discretely-valued and a causal diagram is available, the causal inference problem can be written in its most general form as a polynomial program, due to \citet{balke1994bounds,balke1997bounds} and expanded to general causal diagrams by \citet{zhang2022partial,duarte2024automated}. Different assumptions, see e.g. \citet{rosenbaum2002observational,tan2006distributional,masten2018identification}, may then be interpreted as equality and inequality constraints on this program, and in principle solved exactly with optimization solvers giving a tight interval for the causal query of interest under a remarkably varied set of assumptions. In practice, however, the exponential increase in the number of parameters and complexity of the program render the problem often intractable \citep{shridharan2023scalable}. It is desirable to understand when this optimization program is linear in its free parameters.

In this paper we characterize when that program is linear in its free parameters, for an input diagram in which each confounded component has at most two unobserved confounders. We show that,

\begin{center}
\begin{minipage}{0.85\textwidth}
    \centering
    \textit{Sharp bounds on a counterfactual query given observational probabilities equal the optimum of one linear program if and only if the diagram and the query satisfy a set of conditions that can be read off from the causal diagram.}
\end{minipage}
\end{center}

This result extends the sufficient conditions presented by \citet{symbolicbounds2023general} that already defined a general set of problems that could be solved in a computationally efficient manner. 

We implement this result as a new software package, named \pid{}, that aims to bring together the different families of assumptions that have been proposed in the literature in a unified framework. The user sets the causal or counterfactual query of interest, a variable number of assumptions that can be composed, and provides the observational data. The package then returns a tight numerical interval for the causal query of interest, using our Theorem if applicable, or through bi-linear or polynomial optimization programs for more general cases. This is illustrated in the following example:
\begin{lstlisting}
import pandas as pd
from partialid import Assumptions, Query, bound, from_dataframe

df = pd.read_csv("notebooks/data/job_training.csv")
d = from_dataframe(df, columns=["treated", "employed"])
Q = Query.interventional(outcome="employed", do={"treated": 1}, value=1)
a = (Assumptions.order("treated", "employed")
     & Assumptions.mtr("employed", "treated"))
r = bound(d.variables, d.P, Q, a)
print(f"{r.lower:.3f} {r.upper:.3f} {r.tight}")   # 0.557 0.857 True
\end{lstlisting}

\section{Background}
\label{sec:setup}

The basic framework of our analysis rests on \textit{structural causal models} (SCMs) \citep[Def. 7.1.1]{pearl2009causality}, which formalize the notion of a data-generating process. In addition to specifying the distribution of data, these models also specify the generative mechanisms that produce them. For the purpose of causal inference and learning, SCMs provide a broad, fine-grained hypothesis space.

An SCM $M$ is a tuple $\langle \V, \U, \1F, P \rangle$ where $\V$ is a set of endogenous variables and $\U$ is a set of exogenous variables. $\1F$ is a set of functions where each $f_V \in \1F$ decides values of an endogenous variable $V \in \V$ taking as argument a combination of other variables in the system. That is, $V \leftarrow f_{V}(PA_V, \U_V)$ where observed parents $PA_V \subseteq \V$ and unobserved parents $\U_V \subseteq \U$. An intervention $do(x)$ replaces the causal mechanism $f_{X}$ with the assignment $X \leftarrow x$ corresponding to a sub-model $\M_x$. We will use the counterfactual notation $Y_x(\u)$ to denote the value of the outcome $Y$ in $\M_x$ given $\U=\u$. Drawing values of exogenous variables $\U$ following the distribution $P(\U)$ induces a counterfactual variable $Y_x$. Specifically, the event $\Y_\x = \y$ (for short, $\y_\x$) can be read as ``$\Y$ would be $\y$ had $\X$ been $\x$''.

With a given SCM $M$, for subsets $\Y, \dots ,\W, \X, \dots, \Z \subseteq \V$, the distribution over counterfactuals $\Y_\x, \dots ,\W_\z$ is defined as:
\begin{equation}
  \label{eq:counterfactual_probability}
  P_M(\y_\x, \dots,  \w_\z) = \int \I\{\Y_\x(\u) = \y, \dots,  \W_\z(\u) = \w \} \ dP(\u).
\end{equation}
The problem formulation will be presented in the language of counterfactuals to encompass any causal or counterfactual statement of interest. For example, $P(\y_\x)$, the causal effect of $\{\X=\x\}$ on $\Y=\y$, is here considered a "single event" counterfactual $\{\y_\x\}$.

Each SCM $M$ can be associated with a \textit{causal diagram} $\G$, where nodes represent endogenous variables $\V$, directed edges represent the arguments $PA_V$ of each function $f_{V}$, and \emph{latent hyperedges} $h\subseteq\V$ represent exogenous sources shared by every variable in $h$ (a bi-directed edge $A\leftrightarrow B$ is the pair $\{A,B\}$; distinct hyperedges carry mutually independent exogenous variables, and unconfounded variables carry independent private noises). We leverage the clustering of nodes in $\G$ into \emph{confounded components} (or $c$-components) from \citep{tian2002general}: the maximal subsets $\C \subseteq \V$ connected through the latent hyperedges of $\G$.
  
\begin{definition}[Counterfactual query]
A counterfactual query is a functional of the probabilities of counterfactual events.
\end{definition}
  
For example, $\{Y_x=y,\,X=x'\}$, $\{Y_{X=x,M_x=m}=y\}$, and $\{Z=z\}$ for $X,Y,Z,M\in\V$ and $x' \in \supp_X,y\in\supp_Y,z\in\supp_Z,m\in\supp_M$ are counterfactual events for which an SCM defines a probability assignment. Queries may be linear combinations of counterfactual events, such as the average treatment effect $P(y_{x})- P(y_{x'})$.

\begin{definition}[Linear query]
\label{def:linear-query}
A query $Q$ is \emph{linear} if it is a linear combination of counterfactual events, or such a combination divided by an observational probability fixed by $P$.
\end{definition}

Throughout this paper, we assume that domains of endogenous variables $\V$ are \textbf{discrete and finite}; exogenous variables $\U$ could a priori take values in any (continuous) domain. The counterfactual probability $P(\Y_\x, \ldots, \Z_\w)$ defined in \Cref{eq:counterfactual_probability} is thus a categorical distribution. We use $\supp_V$ to denote the support of the distribution of $V \in \V$, and assume positivity: $P(\v)>0$ for every configuration $\v\in\supp_\V$.

\subsection{Discrete Causal Models}
In general systems, the set of functions $\{f_V:V\in\V\}$ between variables and probability measures $P(\U)$ that define an SCM $M$ may be arbitrary.

For discrete observables $\V$, however, any set of counterfactuals derived from an SCM $M$ can be equivalently obtained from a canonical SCM $N$ parameterized by categorical probability mass functions $P(U), U\in\U$ with a finite and fixed cardinality $|\supp_U|, U\in\U$. This parameterization is known as the family of \emph{canonical} SCMs proposed in \citep[Def. 2.3]{zhang2022partial}.

\begin{definition}[Canonical SCM]
    \label{def:canonical_scm}
    A canonical SCM is an SCM $N=\langle \U, \V, F,P(\u)\rangle$ defined as follows. The set of endogenous variables $\V$ is discrete. The set of exogenous variables $\U =\{R_V : V \in\V \}$, where $\supp_{R_V} = \{1, \dots , m_V\}$ and $m_V = |\{h_V : \supp_{PA_V} \to \supp_V \}|$. For each $V\in\V$, $f_V\in F$ is defined as $f_V(pa_V, u_V) = h_V^{(r_V)}(pa_V)$.
    % A canonical SCM $M$ is an SCM parameterized by categorical probability mass functions $P(U), U\in\U$ with a finite and fixed cardinality,
    % \begin{align}
    %     |\supp_U| = \prod_{V \in \3C(U)} |\supp_{PA_V} \mapsto \supp_V|.
    % \end{align}
    % $|\supp_{PA_V} \mapsto \supp_V |$ denotes the number of functions from the (discretely-valued) parents of $V$ to the values of $V$. $\3C(U)$ denotes the $c$-component of $U$.
\end{definition}

The canonical parameterization is maximally expressive, that is, it can represent \emph{any} SCM $M$ with discrete observables $\V$ \citep[Thm. 2.4]{zhang2022partial}. Let $\R=\{R_V:V\in\V\}$ collect all exogenous variables in the canonical parameterization of $M$. Any counterfactual event can then be evaluated by the weighted sum of exogenous variables whose evaluation lands in that event:
\begin{equation}
  \label{eq:counterfactual_event_probability}
  P_M(\y_\x, \dots,  \w_\z) = \sum_{\r\,:\,\Y_\x(\r) = \y, \dots,  \W_\z(\r) = \w } P_N(\r).
\end{equation}
On a causal diagram $\G$ with latent hyperedges, a canonical mechanism induces $P_N(\r)$ by expressing each $R_V$ as a function of the independent latents $(U_h)_{h\ni V}$ (or of a private noise when no hyperedge contains $V$), so distinct $c$-components factor as $P_N(\r)=\prod_C q_C(\r_C)$.

\begin{example}[Binary bow graph]
\label{ex:bow}
Let $\1G:\{X \rightarrow Y, X \leftrightarrow Y\}$ with binary $X,Y \in \{0,1\}$. An SCM corresponding to this structure parameterized by dependent latent variables $R_X,R_Y$ with cardinality $|\supp_{R_X}| = 2, |\supp_{R_Y}| = 4$ is a canonical SCM. For illustration, encode $R_Y$ as pairs according to the assignment of $f_Y$; for example, $r_Y=(0,0)$ refers to the assignment: $h_Y^{(r_Y)}(x=0)=0, h_Y^{(r_Y)}(x=1)=0$, while $r_Y=(0,1)$ refers to the assignment: $h_Y^{(r_Y)}(x=0)=0, h_Y^{(r_Y)}(x=1)=1$. The joint $\R$ space has $2\times 4=8$ points that fully define all probabilities implied by the SCM. In this particular example, $P(\r)$ can be interpreted as defining the joint counterfactual probability $P(Y_0{=}a, Y_1{=}b, X{=}x)$, which (together with the graph) is sufficient to derive any other observational, interventional, or counterfactual probability. The canonical parameterization extends this assignment to any input causal diagram, guaranteeing that $P(\r)$ is expressive enough to model any counterfactual probability (compatible with the causal diagram).\qed

% \begin{table}[t]
% \centering
% \caption{Canonical parameterization of the binary bow. The factual cell is given by $(X,Y)=(R_X,Y_{R_X})$.}
% \label{tab:bow-canonical}
% \small
% \begin{tabular}{@{}cccccc@{}}
% \toprule
% $(R_X,R_Y)$ & $X$ & $Y_0$ & $Y_1$ & Factual cell & In $P(Y_1{=}1)$? \\
% \midrule
% $(0,00)$ & $0$ & $0$ & $0$ & $(0,0)$ & no \\
% $(0,01)$ & $0$ & $0$ & $1$ & $(0,0)$ & yes \\
% $(0,10)$ & $0$ & $1$ & $0$ & $(0,1)$ & no \\
% $(0,11)$ & $0$ & $1$ & $1$ & $(0,1)$ & yes \\
% $(1,00)$ & $1$ & $0$ & $0$ & $(1,0)$ & no \\
% $(1,01)$ & $1$ & $0$ & $1$ & $(1,1)$ & yes \\
% $(1,10)$ & $1$ & $1$ & $0$ & $(1,0)$ & no \\
% $(1,11)$ & $1$ & $1$ & $1$ & $(1,1)$ & yes \\
% \bottomrule
% \end{tabular}
% \end{table}

\end{example}

\subsection{The (partial) identification program}
\label{sec:program}

The partial identification problem from observational data may then be written as follows.

\begin{definition}[Optimal Counterfactual Bound]
  \label{def:optimal_bound}
  For a distribution $P(\V)$ and assumptions $\1A$, the optimal bound over a counterfactual probability $P(\y_\x, \ldots, \z_\w)$ is defined as, respectively,
  \begin{align}
    \label{eq:counterfactual_bounds}
    \min/\max_{M} \quad P_M(\y_\x, \ldots, \z_\w) \quad\text{s.t.}\quad M \in \{M: P_M(\V) = P(\V),\ M\in\1A\}.
  \end{align}
\end{definition}

Here, the set of assumptions $\1A$ is a restriction on the structural models $M$ under consideration that is assumed to hold in a particular application. They specify additional functional or distributional properties of the SCM. 

%For example, assuming a causal diagram $\G$ enforces that the underlying data generating system $M \in \1A:=\{M:\G_M=\G\}$ induces $\G$. So called shape restrictions, typically in addition to structural restrictions, constrain the shape of functional dependencies. For example, one might enforce that the effect of treatment necessarily increases the probability of outcome enforces $M\in\1A := \{M: P_M(Y_{X=1} = 1) > P_M(Y_{X=0} = 1)\}$, also known as the \emph{monotone treatment response} (MTR) assumption \citep{manski1997monotone}. One may also consider sensitivity assumptions \citep{tan2006distributional,rosenbaum2002observational,masten2018identification,jin2023sensitivity}, that instead cap the strength of (typically) unobserved confounding by an amount specified by the user.

\begin{example}[\Cref{ex:bow} continued]
  Consider the query $P(Y_1{=}1)$. Using the canonical parameterization, we can enumerate the realizations of $\R$ that satisfy the counterfactual event $\{Y_1{=}1\}$. If we take the interpretation of $P(\r)$ as parameterizing the joint $q_{x,ab}=P(R_X{=}x,\,R_Y{=}ab)=P(X=x,Y_0=a,Y_1=b)$, where $x,a,b \in \{0,1\}$, we obtain $P(Y_{1}{=}1)=q_{0,01}+q_{0,11}+q_{1,01}+q_{1,11}$ by marginalizing over the support of $\{X,Y_0\}$. The optimal bounds for this quantity can be obtained from the following program:
  \begin{align}
    \label{eq:bow_bounds}
    \min/\max_{\q} \quad & q_{0,01}+q_{0,11}+q_{1,01}+q_{1,11} \\
    \text{s.t.}\quad & q_{0,00}+q_{0,01}=P(0,0),\quad q_{0,10}+q_{0,11}=P(0,1), \nonumber\\
    & q_{1,00}+q_{1,10}=P(1,0),\quad q_{1,01}+q_{1,11}=P(1,1),\quad q_{x,ab} \ge 0. \nonumber
  \end{align}
  Both the objective and constraints are linear in $\q$.\qed
\end{example}

A linear program is useful here because its optimum is the sharp bound, and a solver can return that optimum. The next section explores under what conditions the partial identification program is a linear program.

\section{Necessary and Sufficient conditions for Linearity}
\label{sec:theorem}
This section provides necessary and sufficient conditions for the partial identification program to be linear in its free parameters, and records which assumptions preserve that linear program.

For a fixed diagram in which every $c$-component has at most two hyperedges, the partial identification program for a linear query (\cref{def:linear-query}) can be reduced to a simpler program that keeps one response law for each $c$-component the query uses, with the observational $c$-factors as constraints. A duplicate hyperedge, or a hyperedge contained in another, may be deleted without changing the realizable laws: every variable on the smaller hyperedge already hears the larger one, so both latents can be stored on the larger hyperedge. All hyperedge counts and graphical tests below apply after that deletion. In a causal order of $\G$, the \emph{$c$-factor} of a $c$-component $C$ is $P_C=\prod_{V_k\in C}P(v_k\mid\mathbf{v}_{<k})$ \citep{tian2002general}, and it is \emph{proper} when it depends on $\mathbf{v}$ only through $C$ and its outside parents $\Pa(C)\setminus C$.

\subsection{The reduced program}
\label{sec:reduced}

\begin{definition}[Reduced program]
  \label{def:reduced}
  Let $\G$ be a directed acyclic graph with latent hyperedges, fix a causal order of $\G$, and let $Q$ be a linear query (\cref{def:linear-query}). An event \emph{reads} a $c$-component when its value depends on the response type of some variable in the component. That variable is found by recursively expanding potential outcomes along directed ancestors in the graph with arrows into the intervened variables removed. A $c$-component is \emph{removed} when no event of $Q$ reads it, or every event that reads it depends on it only through observed roots; the query then uses the corresponding margin of $P$. An unconfounded variable is \emph{pinned} when every event of $Q$ (after that recursive expansion) evaluates it in at most one world, and it enters $Q$ through the kernel $\kappa(v\mid\mathrm{pa})=P(V{=}v\mid\Pa_V{=}\mathrm{pa})$.

  Let $C_1,\ldots,C_K$ be the $c$-components neither removed nor pinned, let $f(q_1,\ldots,q_K)$ be $Q$ after substituting those kernels and root margins, and let $F_j(P)$ be the set of realizable response-type laws $q_j$ of $C_j$ satisfying $A_j q_j=P_j$, where $P_j$ is the $c$-factor of $C_j$. The \emph{reduced program} minimizes and maximizes $f(q_1,\ldots,q_K)$ over $(q_1,\ldots,q_K)\in F_1(P)\times\cdots\times F_K(P)$ (reducing to the substituted functional of $P$ when $K=0$), and is infeasible when a pinned factor is improper, $P(V\mid\mathbf{v}_{<V})\neq P(V\mid\mathrm{pa}_V)$, or a removed component has an improper $c$-factor.

  When $K\ge 1$, the reduced program reduces to the single linear program
  \begin{equation}
  \label{eq:lp}
  \min/\max\sum_{j=1}^{K} c_j^\top q_j
  \quad\text{s.t.}\quad
  A_j q_j = P_j,\quad \mathbf{1}^\top q_j=1,\quad q_j\ge 0,
  \end{equation}
  (where a pinned kernel may appear in $c_j$), paired—whenever a remaining component $C_j$ has two hyperedges $H_1,H_2$ and $H_2$ contains every variable in $C_j$ that has a parent, with \emph{root block} $S_1=C_j\setminus H_2$, $S_2=C_j\setminus H_1$, and $S_{12}=H_1\cap H_2$—with
  \begin{equation}
  \label{eq:two}
  \sum_{r_{12}}q_j(r_1,r_{12},r_2)
  =
  P(S_1{=}r_1)\sum_{r_1',r_{12}}q_j(r_1',r_{12},r_2)
  \quad\text{for all }r_1,r_2,
  \end{equation}
  precisely when each $F_j(P)$ equals that polytope and $f(q_1,\ldots,q_K)=\sum_{j=1}^K c_j^\top q_j$ on $F_1(P)\times\cdots\times F_K(P)$.
\end{definition}

\begin{definition}[Saturation]
  \label{def:saturation}
  A removed $c$-component $C$ is \emph{saturated} when it realizes every proper conditional kernel on $C$ given $\Pa(C)\setminus C$ (equivalently, every proper $c$-factor arising in any causal order of $\G$). An unconfounded variable is always saturated: every proper factor is the kernel $P(V\mid\mathrm{pa}_V)$.

  For a $c$-component $C$, \emph{reachability} holds when, for every variable $Y\in C$ and every parent $Z$ of $C$ outside $C$, either $Z$ is a parent of $Y$, or a directed path runs from $Y$ to $Z$. One hyperedge is saturated exactly when reachability holds.

  For two hyperedges, write $L$ and $R$ for the variables private to each and $I$ for the variables on both. The \emph{cross-edge test} holds when every variable of $L$ and every variable of $R$ are joined by a direct edge.

  A private variable $W$ is \emph{blind} to an intersection variable $A$ when some outside parent $Z$ is a parent of $W$, is not a parent of $A$, and $A$ is not a parent of $W$. \emph{Two-sided blindness} holds when some intersection variable has a blind private variable on both sides. Two hyperedges are saturated exactly when reachability, the cross-edge test, and the absence of two-sided blindness all hold. With no outside parent, reachability holds and no private variable is blind, so the cross-edge test is the whole condition.
\end{definition}

\begin{example}
  On $\G:\{A\to Z\to M,\, Z\to Y,\, M\to Y,\, A\leftrightarrow M,\, A\leftrightarrow Y\}$, the hyperedges are $\{A,M\}$ and $\{A,Y\}$. The variable $A$ lies on both, $M$ lies only on $\{A,M\}$, and $Y$ lies only on $\{A,Y\}$. There is no directed edge from $A$ to $M$ or from $A$ to $Y$. The variable $Z$ lies on no hyperedge: it is a parent of $M$ and of $Y$, and it is not a parent of $A$, so $M$ and $Y$ are both blind to $A$. The table in which $A$ and $Z$ are independent and each equal to $1$ with probability $\tfrac12$, $Y=A$, and the two conditional gaps of $M$ given $A$ equal $0.7$ and $0.6$, is not reproducible (\cref{app:saturation}).\qed
\end{example}

\begin{example}
  \label{ex:instrument}
  On the instrument $\G:\{Z\to X\to Y,\,X\leftrightarrow Y\}$ with binary $Z,X,Y$, the $c$-components are $\{Z\}$ and $\{X,Y\}$. The event $\{Y_x{=}y\}$ sets $X$ to $x$ and evaluates $Y$,
  \begin{equation}
  P(Y_x{=}y)=\sum_{r_X,r_Y}\I\{r_Y(x){=}y\}\,P(r_X,r_Y).
  \end{equation}
  It does not depend on $r_Z$, so it reads $\{X,Y\}$ only and $\{Z\}$ is removed. The event $\{Y_x{=}y,\,Z{=}z\}$ reads $\{Z\}$ in the factual world. The variable $Z$ lies on no hyperedge and is evaluated once, so it is pinned,
  \begin{equation}
  P(Y_x{=}y,\,Z{=}z)
  =
  P(Z{=}z)\sum_{r_X,r_Y}\I\{r_Y(x){=}y\}\,P(r_X,r_Y),
  \end{equation}
  and $P(Z{=}z)$ is a coordinate of the table. A component read only through a root is removed in the same way. On $\{A\to B,\,A\to X\to Y,\,A\leftrightarrow B,\,X\leftrightarrow Y\}$ with query $P(Y_{X=1}{=}1)$, the component $\{A,B\}$ is read only through the root $A$ and is removed, while $\{X,Y\}$ remains.

  The query $P(Z{=}1)$ removes $\{X,Y\}$, and that component is not saturated. Reachability fails: $Z$ is a parent of $X$ and not of $Y$, and no directed path runs from $Y$ to $Z$. The assignments
  \begin{align*}
  Z = f_Z(U_Z), \
  X = R_X(Z,U), \
  Y = R_Y(X,U), \
  \end{align*}
  together with $R_X(z,U)=1$, $P(Y{=}1\mid Z{=}0)=1$, and $P(Y{=}1\mid Z{=}1)=0$ give a proper factor of $\{X,Y\}$, and $Z$ is binary with $P(Z{=}1)=\tfrac12$. Then $Y=R_Y(1,U)$ and $U\perp Z$, so $Y\perp Z$, and $P(Y{=}1,Z{=}0)=\tfrac12$ against $P(Y{=}1)P(Z{=}0)=\tfrac14$. A program that returns $P(Z{=}1)$ on this table reports a number the diagram does not attain. On $Y\to Z\to X$ with the same hyperedge $\{X,Y\}$, $Y$ reaches $Z$ and reachability holds.\qed
\end{example}

\begin{definition}[Covering]
  \label{def:covering}
  A remaining $c$-component is \emph{covered} when it is an unconfounded variable, or one hyperedge, or two hyperedges with a root block (\cref{def:reduced}).
\end{definition}

\begin{example}
  On $\G:\{X\to W\to Y,\, X\leftrightarrow W,\, W\leftrightarrow Y\}$, the variable $X$ lies only on $\{X,W\}$ and $Y$ lies only on $\{W,Y\}$, and there is no edge between $X$ and $Y$, so the cross-edge test fails. If the component is removed it is not saturated. The equations are $X=f_X(U)$, $W=f_W(X,U,V)$, $Y=f_Y(W,V)$, with $U\perp V$. Set $W=0$ and let $X=Y$ with $P(X{=}1)=\tfrac12$. Every parent value that $X$ and $Y$ see is the constant $0$, so $X$ is a function of $U$ and $Y$ is a function of $V$. Independence gives $X\perp Y$, against $P(X{=}1,Y{=}1)=\tfrac12$. Adding the edge $X\to Y$ makes the cross-edge test hold.

  If the component remains, the variables with a parent are $W$ and $Y$, and $\{W,Y\}$ contains both, so $X$ is the root block and covering holds (\cref{prop:lift}). On $\G:\{X\to Y\to D,\, X\leftrightarrow Y,\, X\leftrightarrow D\}$, the variables with a parent are $Y$ and $D$, and neither hyperedge contains both, so there is no root block and covering fails (\cref{ex:desc}).\qed
\end{example}

\begin{definition}[Linearity of the objective]
  \label{def:linearity}
  Write $q_1,\ldots,q_K$ for the response-type laws of the components that remain, and write $f(q_1,\ldots,q_K)$ for the query after pinning and after the root margins of removed components have been substituted. A denominator that is one coordinate of a positive table scales $f$ by a positive constant. The objective is \emph{linear} when, for every table, $f$ is linear on the product of the feasible sets. Equivalently, for every pair of remaining vectors the four-term difference
  \begin{equation}
  \label{eq:four}
  f(q,r')+f(q',r)-f(q,r)-f(q',r')
  \end{equation}
  is zero at all feasible values, the other vectors being held fixed. On each simplex $\mathbf{1}^\top q_j=1$, an affine function of the product is a homogeneous linear function $\sum_j c_j^\top q_j$.

  The difference vanishes for every table exactly when the following reduction leaves no product of two or more remaining vectors. Each counterfactual event in $f$ expands into multilinear monomials in the $q_j$. After collecting terms across all events of $f$, whenever a linear functional of $q_j$ lies in the row space of the equality-constraint matrix of $C_j$ (including~\eqref{eq:two} when $C_j$ has two hyperedges and a root block), replace it by the corresponding linear combination of table entries.
\end{definition}

\begin{example}
\label{ex:linearity}
On $\G:\{X\to Y,\, X\leftrightarrow Y,\, Y\to Z\to W,\, Z\leftrightarrow W\}$, the sum $P(Y_x{=}1)+P(W_z{=}1)$ is two events, each using one remaining vector, and the four-term difference is zero. The conjunction $P(Y_x{=}1,W_z{=}1)$ is the product of the same two factors. The equations of $\{X,Y\}$ do not fix $P(Y_x{=}1)$, so on a positive table each factor takes two values and the difference is nonzero (\cref{ex:two-bows}).

The product $P(Y{=}1,\,W_{Z=z}{=}1)$ uses both vectors and stays linear. On every law of $\{X,Y\}$ that matches the table,
\begin{equation}
\sum_{x,r}\I\{r(x){=}1\}\,q_{XY}(x,r)=P(Y{=}1),
\end{equation}
because the factual outcome equals $R_Y(R_X)$. The objective equals that constant times a linear function of $q_{ZW}$. Replacing the factual $Y$ by $Y_{X=x}$ removes the identity with $P(Y{=}1)$, and both factors move on a positive table.\qed
\end{example}

\subsection{The criterion}
\label{sec:criterion}

\begin{theorem}[Sharp bounds as a linear program]
\label{thm:lp}
Suppose every $c$-component of $\G$ has at most two hyperedges, $Q$ is a linear query (\cref{def:linear-query}), and the response laws are otherwise unrestricted. For every positive table, the reduced program (\cref{def:reduced}) is infeasible precisely when no canonical mechanism reproduces the table, and otherwise its minimum and maximum equal the sharp bounds. This holds if and only if every removed component is saturated, every remaining component is covered, and the objective is linear.
\end{theorem}

The proof is in \Cref{app:saturation}. It uses the pinning reduction (\Cref{lem:middle}), saturation of one and two hyperedges, and the fact that two latents in one $c$-component impose only the independence \Cref{eq:two} when one side consists of roots (\Cref{lem:two}).

\begin{example}
  Consider $\G:\{X\to W\to Y,\, X\leftrightarrow W\}$, with binary $X,W,Y$ and $Q=P(Y_x{=}y)$, and with no further assumptions. The $c$-components are $\{X,W\}$ and $\{Y\}$. The event evaluates $Y$ in one world, so $Y$ is pinned and the only component that remains is $\{X,W\}$. Covering holds because that component has one hyperedge, and linearity holds because the objective is one vector. The reduced program has $K=1$, with $q$ the law of $(R_X,R_W)$ and with $P_1(x',w)=P(x',w)$ the $c$-factor of $\{X,W\}$. The model is compatible only when $P(y\mid x,w)=P(y\mid w)$; the pinned kernel $\kappa(y\mid w)=P(Y{=}y\mid W{=}w)$ then sits in the objective, and the identified interval equals
  \begin{equation}
  \min/\max_{q}
  \sum_{r_X,r_W}\kappa\bigl(y\mid r_W(x)\bigr)\,q(r_X,r_W)
  \end{equation}
  subject to
  \begin{equation}
  \sum_{r_X,r_W}
  \I\{r_X{=}x',\,r_W(x'){=}w\}\,q(r_X,r_W)
  =
  P(x',w)
  \quad\text{for all }x',w,
  \end{equation}
  together with $\sum q=1$ and $q\ge 0$. The first display is $c^\top q$ and the second is $Aq=P_1$.

  For the query $Q=P(Y_{X=1}{=}1,\,Y_{X=0}{=}0)$, the event evaluates $Y$ in two worlds, so $Y$ is not pinned. It reads $\{X,W\}$ and $\{Y\}$, the probability keeps the product $P(r_X,r_W)\,P(r_Y)$, and linearity fails.
\end{example}

\subsection{Linear assumptions}
\label{sec:assumptions}

\begin{definition}[Linear assumption]
\label{def:labels}
An assumption set is \emph{linear} when every constraint is either a zero restriction on a coordinate of a remaining response law $q_j$, or a linear inequality in the remaining vectors $q_1,\ldots,q_K$ whose coefficients are fixed by $P$.
\end{definition}

A zero restriction is a face of the simplex: exclusions, monotone response, no-defiers, and other support constraints \citep{manski1997monotone,imbens1994identification}. An inequality with coefficients fixed by $P$ is an \emph{anchored} inequality. The odds-ratio bounds of \citet{tan2006distributional}, \citet{rosenbaum2002observational}, and \citet{masten2018identification} are of this form once each odds is compared with the observed propensity (\cref{app:msm}). The inequalities may run across remaining components, and every constraint acts on $q_1,\ldots,q_K$.

\begin{proposition}[Linear assumptions]
\label{prop:assumptions}
Suppose every $c$-component has at most two hyperedges and the three tests of \cref{thm:lp} hold. Let the assumption set be linear (\cref{def:labels}). For every positive table, the reduced program with those restrictions adjoined is infeasible when no canonical mechanism obeying the assumptions reproduces the table, and otherwise its minimum and maximum are the sharp bounds under the assumptions.
\end{proposition}

The proof is in \cref{app:thm}.

% \begin{remark}
% \label{rem:assumptions}
% The necessity of the three tests is proved for the diagram and the query alone. An assumption that deletes every non-constant response of a non-root can make that variable behave as a root, so the restricted feasible set can be a polytope on a graph where covering fails (\cref{ex:desc}). An odds comparison that is not anchored is bilinear in $q$ (\cref{ex:gamma}).
% \end{remark}

\begin{corollary}[Solver]
\label{cor:solver}
If the three tests of \cref{thm:lp} hold, the sharp bounds are one LP, or a functional of $P$ when $K=0$. A linear assumption keeps this conclusion (\cref{prop:assumptions}). If one covered component remains and an assumption is an unanchored odds bound (\cref{ex:gamma}), the program has one simplex and bilinear constraints. If linearity fails and two vectors remain, the program is bilinear in those vectors. A product of two vectors in which the factor equations fix one factor remains the linear case. A component with three or more hyperedges lies outside \cref{thm:lp}. Three or more remaining vectors, with a product that the factor equations do not reduce, leave a polynomial program.
\end{corollary}

The degree of the program in the latent laws counts the free latent factors that survive pinning and removal. It is a property of a formulation. A $c$-component with two latents has degree two in the latent laws, and covering makes its feasible set the polytope of \cref{lem:two}.

\section{Worked examples}
\label{sec:illustrations}

Each example is a diagram on which \cref{thm:lp} is a linear program and no bipartition meets Proposition~2 of \citet{symbolicbounds2023general}. The failure of each bipartition is recorded in \cref{app:sachs}. The query is a counterfactual event or a linear combination of such events. The interval returned by \texttt{bound} is the identified range.

\subsection{A pinned mediator}
\label{sec:ill-mediated}
\label{sec:ill-solver}

On $\G:\{X\to W\to Y,\, X\leftrightarrow W\}$ the latent of $\{X,W\}$ is independent of the noise of $Y$. The table is
\begin{equation}
\label{eq:re3}
P(x,w)=\begin{pmatrix}0.25 & 0.25\\ 0.15 & 0.35\end{pmatrix},
\qquad
P(Y{=}1\mid W{=}w)=(0.4,\,0.6),
\end{equation}
with rows indexed by $x$ and $Y\perp X\mid W$. The query is the combination
\[
P(Y_{X=1}{=}1)-P(Y_{X=0}{=}1).
\]
Each event evaluates $Y$ in one world, so $Y$ is pinned to $P(Y\mid W)$ in both, the remaining component $\{X,W\}$ has one hyperedge, and the difference is linear in that one vector. The sharp interval is $[-0.080,0.120]$. The single term $P(Y_{X=1}{=}1)$ is the LP of \cref{ex:mediated}, with interval $[0.470,0.570]$.

$Y$ does not share the latent of $\{X,W\}$, so no bipartition meets their conditions. Replacing the latent by one hyperedge on $\{X,W,Y\}$ drops the independence and widens the combination to $[-0.480,0.520]$. The cross-world event $P(Y_{X=1}{=}1,\,Y_{X=0}{=}0)$ evaluates $Y$ in two worlds, so $Y$ is not pinned and the program is no longer linear.

\subsection{An instrument confounded with the treatment}
\label{sec:ill-designs}

A randomized offer $Z$ encourages enrollment $X$, with outcome $Y$. The law of $(Z,X,Y)$, in that order, is
\begin{equation}
\label{eq:iv-table}
\begin{aligned}
P(0,0,0)&=0.3543,& P(0,0,1)&=0.1173,& P(0,1,0)&=0.2024,& P(0,1,1)&=0.0860,\\
P(1,0,0)&=0.0377,& P(1,0,1)&=0.0421,& P(1,1,0)&=0.0923,& P(1,1,1)&=0.0679.
\end{aligned}
\end{equation}
The query is the joint counterfactual $P(Y_{X=1}{=}1,\,Z{=}0)$. On $\G:\{Z\to X\to Y,\, X\leftrightarrow Y\}$ the instrument lies on no hyperedge and is evaluated once, so it is pinned, and the interval is $[0.215,0.468]$, namely $P(Z{=}0)$ times the Balke--Pearl range $[0.283,0.615]$ for $P(Y_{X=1}{=}1)$; this graph meets their Proposition~2 for the marginal.

On $\G:\{Z\to X\to Y,\, Z\leftrightarrow X,\, X\leftrightarrow Y\}$ the offer shares a latent with treatment and not with the outcome. The variable $Z$ is a root of that component, so the independence of the two latents is the linear constraint of \cref{prop:lift}. Both intervals $[0.215,0.468]$ and $[0.283,0.615]$ persist (whereas one hyperedge on $\{Z,X,Y\}$ drops that independence and widens the joint to $[0.086,0.558]$). An anchored odds-ratio bound of $\Lambda=3$ on the propensity of $X$ adds linear inequalities in that response law whose coefficients are fixed by the table \citep{tan2006distributional}. The program remains linear (\cref{prop:assumptions}), and the joint interval narrows to $[0.251,0.370]$. Every bipartition is met by the latent on $\{Z,X\}$ or by the latent on $\{X,Y\}$, so Proposition~2 does not apply.

\subsection{Two instruments}
\label{sec:ill-two}

On $\G:\{Z_1\to X,\, Z_2\to X,\, X\to Y,\, X\leftrightarrow Y\}$ the instruments are independent. The law of $(Z_1,Z_2,X,Y)$ is
\begin{equation}
\label{eq:two-iv}
\begin{aligned}
P(0,0,0,0)&=0.0159,& P(0,0,0,1)&=0.0062,& P(0,0,1,0)&=0.0059,& P(0,0,1,1)&=0.0069,\\
P(0,1,0,0)&=0.0254,& P(0,1,0,1)&=0.0084,& P(0,1,1,0)&=0.0122,& P(0,1,1,1)&=0.0117,\\
P(1,0,0,0)&=0.0953,& P(1,0,0,1)&=0.0439,& P(1,0,1,0)&=0.0669,& P(1,0,1,1)&=0.1358,\\
P(1,1,0,0)&=0.2643,& P(1,1,0,1)&=0.1175,& P(1,1,1,0)&=0.0711,& P(1,1,1,1)&=0.1126.
\end{aligned}
\end{equation}
The query is the combination
\[
P(Y_{X=1}{=}1,\,Z_1{=}1)+P(Y_{X=0}{=}0,\,Z_2{=}1).
\]
Each event evaluates one instrument in the factual world and the outcome in one world. Both instruments lie on no hyperedge, so each is pinned, and the objective is linear in the response law of $\{X,Y\}$. The sharp interval is $[0.652,1.210]$ (versus $[0.538,1.267]$ under one hyperedge on $\{Z_1,Z_2,X,Y\}$).

The cut with all arrows pointing the same way puts both instruments on the left and $\{X,Y\}$ on the right. Their condition that the left side share one latent fails, since $Z_1\perp Z_2$.

\subsection{A pair chain}
\label{sec:ill-pair}

On $\G:\{X\to W\to Y,\, X\leftrightarrow W,\, W\leftrightarrow Y\}$ the joint law of $(X,W,Y)$ is
\begin{equation}
\label{eq:re2}
\begin{aligned}
P(0,0,0)&=0.20,& P(0,0,1)&=0.05,& P(0,1,0)&=0.10,& P(0,1,1)&=0.15,\\
P(1,0,0)&=0.05,& P(1,0,1)&=0.10,& P(1,1,0)&=0.10,& P(1,1,1)&=0.25,
\end{aligned}
\end{equation}
whose $(X,Y)$-margin is
\begin{equation}
\label{eq:re1}
\begin{aligned}
P(X{=}0,Y{=}0)&=0.30,& P(X{=}0,Y{=}1)&=0.20,\\
P(X{=}1,Y{=}0)&=0.15,& P(X{=}1,Y{=}1)&=0.35.
\end{aligned}
\end{equation}
The variable $X$ is a root, so covering holds and \cref{prop:lift} makes every linear query an LP in the joint response types. The combination $P(Y_{X=1}{=}1)+P(W_{X=1}{=}1)$ has sharp interval $[0.800,1.600]$. The cross-world event $P(Y_{X=1}{=}1,\,Y_{X=0}{=}0)$ has sharp interval $[0.000,0.600]$.

One hyperedge on $\{X,W,Y\}$ meets their Proposition~2. It widens the combination to $[0.750,1.600]$ and the cross-world event to $[0.000,0.650]$. The two-latent graph has no such bipartition: neither latent contains all three variables, and each latent meets both sides of the cut that would separate its endpoints (\cref{ex:pair}).

Monotone response of $W$ in $X$ sets to zero every response of $W$ that decreases in $X$ \citep{manski1997monotone}. The restriction is a face of the remaining response law, so the program stays linear (\cref{prop:assumptions}). The combination narrows to $[1.150,1.600]$ and the cross-world event to $[0.000,0.450]$.

\section{Conclusion}
\label{sec:conclusion}

In this paper, we have shown that for every diagram in which each $c$-component has at most two hyperedges, and across every causal order and positive table, the reduced partial identification program for a linear query reduces to one linear program returning reproducibility and the sharp bounds if and only if the three tests of \cref{thm:lp} hold. Linear assumptions, as in \cref{prop:assumptions}, preserve that conclusion: the reduced program stays a linear program, and the optimum remains the identified set under the assumptions. We introduce a new software package, \pid{}, that reads the tests off the diagram and the query, returns the interval, and records whether the interval is the identified set.

We have also shown that a component with three or more hyperedges lies outside the equivalence. We do not treat continuous variables, symbolic closed forms, sharp product independence, unanchored Rosenbaum $\Gamma$, raw mutual-information budgets, and free-denominator queries at leftover degree two or more. What remains open is a sharp solver for fractional objectives when two simplices still multiply, for independence splits, and for nested optima on every bilinear instance, together with the graphical question for $c$-components with three or more hyperedges.

\section*{Supplementary materials}
The Python package \pid{} and the notebooks that regenerate the
illustrations are available at
\url{https://github.com/alexisbellot/partialid}.
Proofs are collected in \cref{app:proofs}. The package is described in \cref{sec:package}. Related work and the comparisons with existing software are in \cref{sec:related}.

\bibliographystyle{plainnat}
\bibliography{bibliography}

\appendix
\clearpage
\startcontents[appendix]
\printcontents[appendix]{}{1}{%
  \section*{Contents}%
  \setcounter{tocdepth}{1}%
}
\clearpage
\section{Proofs}
\label{app:proofs}

Every nontrivial claim is justified by expanding the query and the
observational law in latents or response types and then eliminating
factors. We write $\ind[\,\cdot\,]$ for the indicator of an event.

\subsection{Canonical partitions}
\label{app:canon}

\begin{proof}[Finite response encoding of \cref{def:canonical_scm}]
Fix $V$ with finite parent set $\Pa_V$. Each noise value $u_V$
determines a function
$r_u:\mathrm{Val}(\Pa_V)\to\mathrm{Val}(V)$ by
$r_u(\mathrm{pa})=f_V(\mathrm{pa},u_V)$. There are only
$|\mathrm{Val}(V)|^{|\mathrm{Val}(\Pa_V)|}$ such functions. Let
$R_V$ enumerate them and set $f_V(\mathrm{pa},r)=r(\mathrm{pa})$.
Then $f_V(\,\cdot\,,u_V)=f_V(\,\cdot\,,r_u)$ by construction, and
$r_u$ is unique. The law of $U_V$ pushes forward to a law on this
finite set. If $\Pa_V=\emptyset$ the functions are the constants in
$\mathrm{Val}(V)$.
\end{proof}

\subsection{Pinning}
\label{app:middle}

\begin{lemma}[Pinning an unconfounded variable]
\label{lem:middle}
Let $V$ be a variable that no hyperedge contains. Suppose every counterfactual event of $Q$
evaluates $V$ in at most one world (the factual world counts as a world), and
$P(\Pa_V=\mathrm{pa})>0$ for every parent configuration $\mathrm{pa}$. Then
every appearance of $V$'s mechanism in the query and in the observational
constraints is through the kernel $\kappa(v\mid\mathrm{pa})=P(V{=}v\mid\Pa_V{=}\mathrm{pa})$.
The observational law is compatible with the model only if
$P(v\mid \mathbf{v}_{<V})=P(v\mid\mathrm{pa}_V)$, where $\mathbf{v}_{<V}$ are
the values of the variables preceding $V$ in a causal order; under that
equality, $V$'s factor can be replaced by $\kappa$.
\end{lemma}

\begin{proof}[Proof of \cref{lem:middle}]
\emph{SCM.} No hyperedge contains $V$, so $V=f_V(\Pa_V,R_V)$ with
$R_V\sim p_V$ independent of the vector $\mathbf{U}'$ of all other latents.
Every other variable, and $\Pa_V$ in particular, is a function of
$\mathbf{U}'$ and of $V$'s value. Because the graph is acyclic, $\Pa_V$ in
any world is a function $\mathrm{pa}(\mathbf{u}')$ of $\mathbf{U}'$ alone, and
so are the variables $\mathbf{V}_{<V}$ that precede $V$ in a causal order.

\emph{Kernel.} Since $R_V\perp\mathbf{V}_{<V}$,
$P(V{=}v\mid\mathbf{v}_{<V})=P(R_V(\mathrm{pa}){=}v)$, where $\mathrm{pa}$ is
the parent part of $\mathbf{v}_{<V}$. By positivity,
$\kappa(v\mid\mathrm{pa}):=P(R_V(\mathrm{pa}){=}v)=P(V{=}v\mid\Pa_V{=}\mathrm{pa})$
is identified, and $P(v\mid\mathbf{v}_{<V})$ depends on $\mathbf{v}_{<V}$ only
through $\mathrm{pa}$. If $P$ violates that equality, no model reproduces $P$.

\emph{Query and data.} Take any counterfactual event $E$ (of $Q$ or an observational cell)
that evaluates $V$ in one world, at parents $\mathrm{pa}(\mathbf{u}')$:
\begin{align*}
P(E)&=\sum_{\mathbf{u}'}q'(\mathbf{u}')\sum_{v}\sum_{r}p_V(r)\,
\ind\bigl[r(\mathrm{pa}(\mathbf{u}'))=v\bigr]\,
\ind[E'(\mathbf{u}',v)]\\
&=\sum_{\mathbf{u}'}q'(\mathbf{u}')\sum_v
\kappa\bigl(v\mid\mathrm{pa}(\mathbf{u}')\bigr)\,
\ind[E'(\mathbf{u}',v)],
\end{align*}
where $E'(\mathbf{u}',v)$ is the event that the remaining literals of $E$
hold when $V$ takes the value $v$ in that world. Each term is linear in $q'$
with identified coefficients.

\emph{Elimination.} $p_V$ enters every counterfactual event only through the identified
one-dimensional margins $\kappa(\cdot\mid\mathrm{pa})$. Substitute them. The
remaining freedom in $p_V$, namely the joint law of $(R_V(\mathrm{pa}))_{\mathrm{pa}}$,
appears nowhere. Conversely, under the equality any $p_V$ with these margins
reproduces the factor $P(v\mid\mathbf{v}_{<V})$ of $P$, so the substitution
does not change what the other factors may do.

If some counterfactual event read $R_V$ in two worlds at $\mathrm{pa}\neq\mathrm{pa}'$, it
would require $P(R_V(\mathrm{pa}){=}v,\allowbreak\,R_V(\mathrm{pa}'){=}v')$, which $P$ does
not identify, so the hypothesis of \cref{lem:middle} would fail.
\end{proof}

\subsection{Classification}
\label{app:thm}

\begin{lemma}[Two latents in one $c$-component]
\label{lem:two}
Let a $c$-component $C$ have exactly two hyperedges, with latents
$U_1\perp U_2$. Write $S_1$ for the variables of $C$ that read only $U_1$,
$S_2$ for those that read only $U_2$, and $S_{12}$ for those that read both.
A joint law $q$ of the response types $(R_{S_1},R_{S_{12}},R_{S_2})$ is
induced by some model if and only if $R_{S_1}\perp R_{S_2}$ under $q$. If
every variable in $S_1$ is a root, the law of $R_{S_1}$ is $P(S_1)$ and the
condition is the linear constraint~\eqref{eq:two} of \cref{thm:lp}.
Given both latents, the response of $S_{12}$ may be any kernel.
\end{lemma}

\begin{proof}[Proof of \cref{lem:two}]
\emph{SCM.} Each variable of $S_1$ reads $U_1$, each variable of $S_2$ reads
$U_2$, and each variable of $S_{12}$ reads both, so
$R_{S_1}=g_1(U_1)$, $R_{S_2}=g_2(U_2)$, and $R_{S_{12}}=g_{12}(U_1,U_2)$.

\emph{Only if.} $R_{S_1}$ is a function of $U_1$ and $R_{S_2}$ of $U_2$, and
$U_1\perp U_2$.

\emph{If.} Take $q$ with $q(r_1,r_2)=q(r_1)q(r_2)$ and set
$\kappa(r_{12}\mid r_1,r_2)=q(r_1,r_{12},r_2)/\bigl(q(r_1)q(r_2)\bigr)$,
with any probability vector where the denominator vanishes. Let
$U_2:=R_{S_2}\sim q(r_2)$. Let $U_1:=(R_{S_1},G)$, where $R_{S_1}\sim q(r_1)$
and, given $R_{S_1}=r_1$, $G$ is a random map from the values of $R_{S_2}$ to
those of $R_{S_{12}}$ whose coordinates $G(r_2)$ are drawn independently from
$\kappa(\cdot\mid r_1,r_2)$. Set $R_{S_{12}}:=G(U_2)$. Then $U_1\perp U_2$,
each response type reads only its own latents, and
\[
P(R_{S_1}{=}r_1,R_{S_{12}}{=}r_{12},R_{S_2}{=}r_2)
=q(r_1)\,q(r_2)\,\kappa(r_{12}\mid r_1,r_2)=q(r_1,r_{12},r_2).
\]

\emph{Roots.} A root has no parents, so its response type is its value and
the law of $R_{S_1}$ is $P(S_1)$. Writing
$q(r_1,r_2)=\sum_{r_{12}}q(r_1,r_{12},r_2)$ and
$q(r_2)=\sum_{r_1',r_{12}}q(r_1',r_{12},r_2)$, the independence is
$q(r_1,r_2)=P(S_1{=}r_1)q(r_2)$, which is~\eqref{eq:two}.
\end{proof}

\subsection{Saturation and covering}
\label{app:saturation}

\begin{lemma}[Agreement]
\label{lem:agree}
Let $X,Y,Z$ be independent, and suppose $f(X,Z)=g(Y,Z)$ almost surely. Then the common value is a function of $Z$ alone.
\end{lemma}

\begin{proof}[Proof of \cref{lem:agree}]
For almost every fixed $z$, the two sides are functions of independent arguments and are therefore constant in those arguments.
\end{proof}

\begin{lemma}[Equal absolute gaps]
\label{lem:gap}
Let $A$, $C$, and $W$ be binary, let $U,V,W$ be independent, and write $A=f(U,V)$ and $C=h(W,U)$. Suppose $0<P(A{=}1)<1$, both values of $W$ occur, and that for each such value $w$ the variable $A$ is almost surely a function of $(h(w,U),V)$. Then one of the following holds. The maps $h(w,\cdot)$ share no bipartition that carries a non-constant binary function, in which case $A$ is a function of $V$ alone, $C\perp A\mid W$, and both gaps are $0$. Or the maps share one bipartition: there is a non-constant bit $\psi(U)$ and a bit $\varepsilon(w)$ with $h(w,U)=\psi(U)\oplus\varepsilon(w)$, and
\[
\bigl|P(C{=}1\mid A{=}1,W{=}w)-P(C{=}1\mid A{=}0,W{=}w)\bigr|
\]
does not depend on $w$.
\end{lemma}

\begin{proof}[Proof of \cref{lem:gap}]
If the maps share no usable bipartition, then for almost every $v$ the map $u\mapsto f(u,v)$ is constant on the level sets of every $h(w,\cdot)$, hence constant, and $A$ is a function of $V$ alone. Since $C=h(W,U)$ does not depend on $V$ and $W\perp(U,V)$, one has $C\perp A\mid W$.

If the maps share one bipartition, $C$ does not depend on $V$ and $W\perp(\psi(U),A)$. Conditioning on $W=w$ flips $\psi$ or leaves it, so the absolute gap of $C$ given $A$ equals the absolute gap of $\psi$ given $A$.
\end{proof}

In the necessity proofs below we exhibit boundary tables in which variables outside a witness set equal $0$ almost surely. Because the canonical response space (\cref{def:canonical_scm}) is finite, the set of tables reproducible on $\G$ is the continuous image of a compact product of simplices and is therefore closed. Any unrealizable boundary table whose factors in a causal order $\prec$ are proper lies at positive distance from that closed set, so mixing each conditional in the chain rule of $\prec$ with a uniform kernel of small weight $\varepsilon>0$ (using only variables in $C_i\cup(\Pa(C_i)\setminus C_i)$ for each $V_k\in C_i$) yields a strictly positive unrealizable table whose $\prec$-factors remain proper. Moreover, any outside parent $Z\notin C$ belongs to a $c$-component $C'\neq C$ sharing no hyperedge with $C$, and $C'$ realizes an independent marginal on $Z$ (with $C'\setminus\{Z\}$ set to $0$) by drawing $Z$ from a seed on a hyperedge meeting $Z$ (or its private noise) and ignoring $\Pa_Z$.

\begin{lemma}[One hyperedge]
\label{lem:reach}
A $c$-component carried by one hyperedge is saturated if and only if reachability holds.
\end{lemma}

\begin{proof}[Proof of \cref{lem:reach}]
\emph{Sufficiency.} Let the factor of $C$ be proper. Give each variable of $C$ a fresh uniform seed on the hidden cause $U$, independent of the exogenous seeds of the unconfounded variables. In a topological order, sample each unconfounded variable from the kernel of its parents, and sample each $V\in C$ from $P(V\mid\text{earlier variables in }C,\,\mathrm{pa}(V))$. This is a function of $\mathrm{pa}(V)$ and $U$: the earlier variables in $C$ are recomputed from their seeds. Every outside parent of $C$ that precedes $V$ in the causal order is a parent of $V$, because otherwise reachability would give a directed path from $V$ to that parent. A proper factor does not depend on a variable $w$ outside $C$ and outside the parents of $C$. Let $V$ be the earliest variable of $C$ whose conditional could depend on $w$. Summing the product over the later variables of $C$ recovers that conditional times the earlier factors. The earlier factors do not depend on $w$, and the sum does not either, so the conditional of $V$ does not. The equations reproduce the chain rule.

\emph{Necessity.} Suppose $Y\in C$, an outside parent $Z$ is not a parent of $Y$, and no directed path runs from $Y$ to $Z$. Let $Z$ be binary with $P(Z{=}1)=\tfrac12$, let $Y=Z$, and let every other variable equal $0$. Every parent of $Y$ equals $0$, so $Y=R_Y(0,U)$. No path carries $U$ from $Y$ to $Z$, and every other variable of $C$ is constant, so $Z\perp U$ and therefore $Y\perp Z$, against $Y=Z$. Because no directed path runs from $Y$ to $Z$, acyclicity admits a causal order $\prec$ placing $Z$ before $Y$; in $\prec$ the component containing $Z$ has proper factor $P(Z{=}1)=\tfrac12$, the dependence $Y=Z$ sits in the proper $\prec$-factor of $C$, and the $\varepsilon$-perturbation above yields a strictly positive impossible table.
\end{proof}

\begin{lemma}[Two hyperedges, no outside parent]
\label{lem:cross}
Let $C$ have two hyperedges and no parent outside $C$. Write $L$ and $R$ for the two private sets. Then $C$ is saturated if and only if every variable of $L$ and every variable of $R$ are joined by a direct edge.
\end{lemma}

\begin{proof}[Proof of \cref{lem:cross}]
\emph{Necessity.} If $A\in L$ and $B\in R$ have no direct edge, the table with $A=B$, $P(A{=}1)=\tfrac12$, and every other variable equal to $0$ is impossible (and its $\varepsilon$-perturbation is a strictly positive impossible table). Every parent of $A$ or of $B$ equals $0$, so $A$ is a function of one latent and $B$ of the other. Independence gives $A\perp B$, against $P(A{=}1,B{=}1)=\tfrac12$.

\emph{Sufficiency, by peeling sources.} The inductive object is a directed graph colored by $L$, $R$, and $I$, with an external input $E$ available to every remaining variable. Variables in $L\cup I$ hear $U$, variables in $R\cup I$ hear $V$, and $U\perp V$. If both private sets are nonempty, every cross pair has a direct edge. Every joint, and every kernel of $E$, is realizable. A component with no outside parent is the case of $E$ empty.

If one private set is empty, every variable hears one common cause. Order the variables, give each a fresh seed on that cause, independent of $E$, and sample the target conditional given the earlier variables and $E$.

If the intersection is empty and both private sets are nonempty, put seeds of $L$ on $U$ and seeds of $R$ on $V$. In a topological order a variable $x\in L$ has every earlier variable of $R$ as a parent, and it recomputes every earlier variable of $L$ from $U$. The symmetric reading applies on $R$.

If $\rho\in L$ has no parent in the remaining graph, the cross edge cannot enter $\rho$, so $\rho$ points at every variable of $R$. Draw $\rho$ from its marginal with a seed on $U$, and apply the induction to the smaller graph with external input $(E,\rho)$. The symmetric construction handles a source in $R$.

If $\rho\in I$ has no parent, delete $\rho$ and every edge out of $\rho$. The remaining graph is smaller and still has a cross edge for every surviving cross pair. Realize its marginal, and reserve a further coordinate of $U$ unused by that mechanism. The source hears both causes, recomputes the rest, and draws itself from the target conditional given the rest and $E$. Children of $\rho$ may ignore the parent $\rho$ in the marginal mechanism; dependence between $\rho$ and its children is carried by the seeds $\rho$ hears.

Every nonempty graph falls under one of these cases, and each reduction removes one variable.
\end{proof}

\begin{lemma}[Two hyperedges with outside parents]
\label{lem:two-sat}
Let $C$ have two hyperedges and at least one outside parent. Then $C$ is saturated if and only if reachability, the cross-edge test, and the absence of two-sided blindness all hold.
\end{lemma}

\begin{proof}[Proof of \cref{lem:two-sat}]
\emph{Reachability and the cross-edge test.} The table of \cref{lem:reach} is impossible whenever some $Y\in C$ misses an outside parent $Z$ and does not reach it: $Y$ is a function of the causes through $Y$, and those causes do not reach $Z$. The table of \cref{lem:cross} is impossible whenever a cross pair has no direct edge, with every outside parent set to $0$.

\emph{Two-sided blindness.} Suppose $m_1$ is private to one hyperedge and blind to an intersection variable $\rho$ via an outside parent $Z_1$, and $y_2$ is private to the other and blind to the same $\rho$ via an outside parent $Z_2$. The cross edge joins $m_1$ and $y_2$. Blindness of $y_2$ says that $\rho$ is not a parent of $y_2$, and blindness of $m_1$ says that $Z_1$ is not a parent of $\rho$. If $m_1\to y_2$, a directed path from $y_2$ back to $Z_1$ would cycle through $m_1$, so reachability puts $Z_1$ among the parents of $y_2$. Then both private variables are blind to $\rho$ via the same parent $Z=Z_1$. The opposite cross edge produces a common parent in the same way. Since $Z$ is not a parent of $\rho$, reachability gives a path from $\rho$ to $Z$, and in every topological order $\rho$ precedes $Z$, which precedes both ends of the cross edge.

Pin every other variable to $0$, let $\rho$ and $Z$ be independent and binary, each equal to $1$ with probability $\tfrac12$, let the head of the cross edge equal $\rho$, and give the tail the conditional gaps $0.7$ and $0.6$. The equations inside $C$ reduce to $A=\rho=f(U,V)$, $M=g(Z,U)$, and $Y=h(M,Z,V)$, with the head playing the role of $Y$ in the smallest picture and the tail the role of $M$. Suppose a mechanism realizes $Y=\rho$ and $Z\perp\rho$. With all variables outside $\{\rho,Z,m_1,y_2\}$ equal to $0$, $Z$ is a function of $\rho$ and latents outside $C$ independent of $(U,V)$, so $Z\perp\rho$ implies $Z\perp(U,V)$. For each value $z$,
\[
f(U,V)=h\bigl(g(z,U),z,V\bigr),
\]
so \cref{lem:gap} applies with conditioning variable $Z$ and inner variable $M$. The absolute gap of $M$ given $\rho$ is the same at both values of $Z$, or both gaps are $0$. The gaps $0.7$ and $0.6$ violate that identity. The factor containing $Z$ is a constant kernel of $\rho$, realized by $Z$'s component ignoring $\rho$, and perturbing by $\varepsilon>0$ yields a strictly positive impossible table.

Equal absolute gaps are realizable, so the identity is sharp. Independent binary $U$ and $Z$, each equal to $1$ with probability $\tfrac12$, a mode bit of weight $0.8$, and the assignments $M=U\oplus Z$ and $\rho=1-U$ when the mode is on, with $\rho$ an independent binary variable equal to $1$ with probability $\tfrac12$ when the mode is off, and with the head copying $\rho$, give gaps $-0.8$ and $+0.8$.

\emph{Sufficiency.} Proceed by induction on $|C|$, and for fixed $|C|$ on the number of outside parents that are not among a set $E$ of common inputs. Reachability, the cross-edge test, and the absence of two-sided blindness are assumed, and every variable may use $E$. If $L$ or $R$ is empty, every variable hears one common cause and \cref{lem:reach} applies. If $I$ is empty, order $C$ topologically, put seeds of $L$ on $U$ and seeds of $R$ on $V$, and sample each variable from its proper conditional: a variable in $L$ has every earlier variable of $R$ as a parent by the cross edge, has every earlier outside parent as a parent by reachability, and recomputes every earlier variable of $L$ from $U$. If every outside parent already lies in $E$, \cref{lem:cross} applies.

Otherwise pick an outside parent $Z\notin E$ earliest in a topological order, and split $C$ into $N=\{v:Z\notin\mathrm{Pa}_v\}$ and $M=\{v:Z\in\mathrm{Pa}_v\}$. If $N$ is empty, move $Z$ into $E$ and apply the induction. If $N$ is nonempty, reachability gives every $v\in N$ a path to $Z$, and acyclicity forces four facts. No edge runs from $M$ to $N$. Every cross edge between $L\cap N$ and $R\cap M$ points into $R\cap M$, and every cross edge between $R\cap N$ and $L\cap M$ points into $L\cap M$. Any outside parent of a variable in $N$ precedes $Z$ and therefore lies in $E$, so a proper factor splits as a kernel of $N$ given $E$ times a kernel of $M$ given $N$, $Z$, and the other outside parents.

No two-sided blindness at $Z$ says that each $\rho\in I\cap N$ fails to be blind on at least one side. If no variable of $R\cap M$ is blind to $\rho$, then $\rho$ is a parent of every such variable; put $\rho$ in a block $I_U$. Otherwise $\rho$ is a parent of every variable of $L\cap M$; put $\rho$ in $I_V$. Realize the kernel of $N$ in three stages. Realize the marginal of the private variables of $N$ by the empty-intersection construction, with seeds of $L\cap N$ on $U$ and seeds of $R\cap N$ on $V$. Order $I_U$ and give each variable a fresh seed on $U$; it hears both causes, recomputes the private variables of $N$, and samples its conditional. Order $I_V$ and give each variable a fresh seed on $V$; it recomputes the private variables and $I_U$.

Every variable of $M$ can then evaluate $N$. A variable in $L\cap M$ hears $U$ and has $R\cap N$, $I_V$, and $E$ as parents. A variable in $R\cap M$ hears $V$ and has $L\cap N$, $I_U$, and $E$ as parents. A variable in $I\cap M$ hears both causes. Thus $E\cup N\cup\{Z\}$ is a common input to $M$. The piece $M$ is smaller and inherits the three conditions: a path used by reachability cannot return from $M$ into $N$, and a two-sided blind pair inside $M$ would already have been two-sided in $C$. The induction realizes the kernel of $M$, and the chain rule gives the factor of $C$.
\end{proof}

\begin{lemma}[Covering]
\label{lem:cover}
Let $C$ be a $c$-component with at most two hyperedges. For every positive table, the realizable response laws compatible with that table form a polytope if and only if $C$ is covered. The polytope may be empty.
\end{lemma}

\begin{proof}[Proof of \cref{lem:cover}]
An unconfounded variable has every law of its response type realizable, and the table cuts that law by the linear kernel equations $\sum_{r:\,r(\mathrm{pa})=v}q(r)=P(V{=}v\mid\mathrm{pa})$. One hyperedge has every response law realizable, because the hidden cause can carry the whole response tuple and every variable hears it. The feasible set is the linear system given by the $c$-factor, together with $q\ge 0$ and $\sum q=1$. It is empty exactly when no response law induces the factor.

Now let $C$ have two hyperedges, and suppose the hyperedge with latent $U$ contains every variable of $C$ that has a parent. Let $S$ be the variables private to the other hyperedge. Each of them is a global root, and the block $S$ is a function of the other latent. By \cref{lem:two}, a joint response law is realizable if and only if the response types private to $U$ are independent of $S$. The table fixes the margin of $S$, so the independence is~\eqref{eq:two}. Together with the linear factor equations this is a polytope.

Conversely, suppose at least two variables of $C$ have a parent and neither hyperedge contains all of them. Then two variables with a parent, $Y$ and $D$, share no hyperedge: a variable of $C$ outside one hyperedge is private to the other, and the two witnesses of the failed containments are distinct. Any realizable law has $R_Y\perp R_D$. Start from a full-support product law $q_0$, which is realizable by drawing each response from a latent that variable hears. For a binary variable the signed measure
\[
\delta=\varepsilon\bigl(\delta_{\mathrm{const}_0}+\delta_{\mathrm{const}_1}-\delta_{\mathrm{id}}-\delta_{\mathrm{neg}}\bigr)
\]
has total mass zero. The pair of constants, and the pair consisting of the identity and the negation, each put mass $1/2$ on both outputs at every parent value, so $\delta$ contributes $0$ to every one-world output margin. Choose $\varepsilon>0$ small enough that $q_0+\delta$ stays nonnegative.

Let $q$ add $\delta$ to the marginal of $R_Y$ only, and let $q'$ add $\delta$ to the marginal of $R_D$ only. Both are product laws, hence realizable. A data cell depends on $R_Y$ only through the output of $Y$ at the parent value fixed by the other responses, and parents are earlier in the topological order. Since $\delta$ preserves every one-world output margin, $q$ and $q'$ reproduce the same strictly positive table. Their midpoint has joint law of $(R_Y,R_D)$ equal to the product of its margins minus $\tfrac14\,\delta\otimes\delta$. The two response types are dependent, so the midpoint is not realizable. The map from response laws to tables is linear, so the midpoint matches the table. The feasible set is not convex.
\end{proof}

\begin{proof}[Proof of \cref{thm:lp}]
\emph{Step 1: one block per $c$-component.} Let $C_1,\ldots,C_m$ be all the
$c$-components and $q_{C}$ the joint law of the response types of the
variables in $C$. Distinct $c$-components share no latent, so the joint law
of all response types is $q_{C_1}\otimes\cdots\otimes q_{C_m}$, and the
model restricts each $q_C$ separately. By the $c$-component factorization
\citep{tian2002general},
\[
P(\mathbf{v})=\prod_{i=1}^m Q[C_i](\mathbf{v}_{C_i}\mid\mathrm{pa}(C_i)),
\qquad
Q[C_i](\mathbf{v}_{C_i}\mid\mathrm{pa})=\sum_{\mathbf{r}}
\ind[g_i(\mathbf{r},\mathrm{pa})=\mathbf{v}_{C_i}]\,q_{C_i}(\mathbf{r}),
\]
where $g_i$ evaluates the types of $C_i$ given the values $\mathrm{pa}$ of
its parents outside $C_i$. Each $c$-factor is identified
from $P$ as a product of conditionals $P(v_k\mid\mathbf{v}_{<k})$ over the
variables of $C_i$ \citep{tian2002general}; call it $P_i$. A model reproduces $P$ only if $A_i q_{C_i}=P_i$ for every $i$, with $A_i$ the
fixed $0/1$ matrix of the display, and only if each factor $P_i$ is a function of the parents of $C_i$. The constraints separate by $c$-component: the feasible set is a product of one block
per $c$-component.

\emph{Step 2: pinning and removal.} A counterfactual event of $Q$ depends only on the blocks it reads. By \cref{lem:middle}, a pinned unconfounded variable enters every such event through the kernel $\kappa$, and the model reproduces $P$ only when that factor is proper. A component removed by \cref{def:reduced} enters the objective through observed root margins, or not at all. Its own factor still has to be proper, and, when it is proper, realizable. If $K=0$, every event is a functional of those kernels and root margins.

\emph{Step 3: sufficiency.} Fix any causal order of $\G$ and any positive table, and suppose the three tests hold. An improper factor of a pinned or removed component makes the table impossible; the reduced program records that and is infeasible. A proper factor of a removed component is realizable by saturation: an unconfounded variable by \cref{lem:middle}, one hyperedge by \cref{lem:reach}, and two hyperedges by \cref{lem:cross,lem:two-sat}. Pinning and the root-margin substitution keep the same set of values of the query. By \cref{lem:cover}, each remaining feasible set $F_j(P)$ is a polytope, cut out by the $c$-factor equations and, when the component has two hyperedges, by~\eqref{eq:two}. The feasible set of the reduced program is the product of these polytopes. Linearity says that the objective is linear on that product, hence of the form~\eqref{eq:lp}. If some remaining polytope is empty, no canonical mechanism reproduces the table. If every remaining polytope is nonempty, the removed components have realizable laws, so some mechanism reproduces the table, and the optimum is the sharp bound.

\emph{Step 4: necessity.} Reproducibility factors by components. Suppose a removed component $C$ fails saturation, missing a proper $\prec$-factor $p_C$ in some causal order $\prec$. Build a table whose $C$-piece in $\prec$ is $p_C$ and whose other pieces come from realizable laws; a point mass on a constant response is realizable. When the missing piece needs an outside parent to take more than one value, give that parent's component a law that charges those values independently of its parents, and perturb by $\varepsilon>0$ to make the table strictly positive while keeping all $\prec$-factors proper. The resulting positive table is not reproducible. The reduced program in $\prec$ does not constrain $C$, and the remaining feasible sets are nonempty, so the reduced program has an optimum on a table the diagram does not produce.

If some remaining component fails covering, \cref{lem:cover} supplies a positive table whose realizable set $F_j(P)$ is not a polytope, so the reduced program's feasible set on that table is not the polytope of~\eqref{eq:lp}--\eqref{eq:two}. If the feasible sets are polytopes and a four-term difference~\eqref{eq:four} is nonzero at feasible values, the objective is not linear on the product. For three or more vectors the difference is tested one pair at a time: a product $q_1 q_3$ has a zero four-term difference in the slots $(q_1,q_2)$ and is still bilinear.
\end{proof}

\begin{proof}[Proof of \cref{prop:assumptions}]
The hypotheses of \cref{thm:lp} hold, so Step~3 of its proof applies. For a positive table, each removed component with a proper factor is realizable, each remaining feasible set is a polytope, and on that product the query equals $\sum_j c_j^\top q_j$. A linear assumption (\cref{def:labels}) does not change the query. Every constraint is $q_{j,i}=0$ or $b(P)^\top q\le d(P)$, where $q=(q_1,\ldots,q_K)$ collects the remaining vectors and the coefficients $b(P),d(P)$ are computed from the table.

\emph{Face.} On the bow $X=f_X(U)$, $Y=f_Y(X,U)$, with $Q=P(Y_1{=}1)$,
\begin{align*}
P(Y_1{=}1)
&=
\sum_r \ind[r_Y(1){=}1]\,q(r),
\\
P(x,y)
&=
\sum_r \ind[r_X{=}x]\,\ind[r_Y(x){=}y]\,q(r).
\end{align*}
Monotone response sets $q(r)=0$ whenever $r_Y(1)<r_Y(0)$. The objective and the factor equations remain linear in the surviving coordinates, and the feasible set is a face of the polytope of \cref{lem:cover}. The resulting interval is~\eqref{eq:mtr}.

\emph{Anchored inequality.} The same bow with an anchored odds bound compares each response with the observed propensity $\pi_0$. Clearing the constant odds, as in \cref{app:msm}, gives
\[
q(x,r)\,\pi_0(x')
\le
\Lambda\,\pi_0(x)\,q(x',r),
\]
linear in $q$. Anchoring replaces the product of two free coordinates in \cref{ex:gamma} by coefficients computed from $\pi_0$.

\emph{Cut.} The restricted feasible set is the product of the remaining polytopes cut by finitely many such equalities and inequalities, including inequalities that run across components. The cut of a polytope by linear constraints is a polytope, and it may be empty. For one hyperedge, and for an unconfounded variable, every response law that matches the $c$-factor is realizable, so a point of the cut is realizable. For two hyperedges with a root block, \cref{lem:two} states realizability as~\eqref{eq:two}, already present in the polytope. A point that meets the factor equations, \eqref{eq:two}, and the assumption is induced by a canonical mechanism. If the cut is empty, no such mechanism reproduces the table. If it is nonempty, the linear objective attains its minimum and maximum on the cut, and those values are the sharp bounds under the assumptions.
\end{proof}

\paragraph{Non-necessity under linear assumptions.}
Necessity of the three tests is Step~4 of the proof of \cref{thm:lp}, which uses no assumption beyond the diagram and the query. An assumption can restore a polytope where covering fails. Let $C$ have two hyperedges, with private blocks $S_1$ and $S_2$ and intersection $S_{12}$, and suppose a variable of $S_2$ has a parent, so $S_2$ is not a root block. Set $q(r_2)=0$ unless every coordinate of $r_2$ is a constant response. Each variable of $S_2$ then ignores its parents, and $S_2$ equals its response. The factor equations are realizable on this face only when the conditional law of $S_2$ does not depend on its parents; otherwise the face is empty. When they are realizable, summing them yields $P(S_2{=}r_2)=\sum_{r_1,r_{12}}q(r_1,r_{12},r_2)$ for each constant $r_2$. \Cref{lem:two} requires $R_{S_1}\perp R_{S_2}$. Substituting the margin, as in the root case of that lemma, gives
\begin{equation}
\label{eq:const-face}
\sum_{r_{12}} q(r_1,r_{12},r_2)
=
P(S_2{=}r_2)\sum_{r_2',r_{12}} q(r_1,r_{12},r_2')
\end{equation}
for every constant $r_2$. This is \eqref{eq:two} with the root block on the $S_2$ side, and the restricted feasible set is a polytope. On $X\to Y\to D$ with hyperedges $\{X,Y\}$ and $\{X,D\}$ (\cref{ex:desc}), $D$ is the non-root private variable, and deleting its non-constant responses is this face.

\begin{proof}[Proof of \cref{cor:solver}]
The first two sentences are \cref{thm:lp} and \cref{prop:assumptions}. An unanchored odds bound on one simplex is the product of two coordinates of that simplex (\cref{ex:gamma}), so the constraints are bilinear while the feasible set stays a polytope. If~\eqref{eq:four} is nonzero and exactly two vectors remain, the objective is bilinear in those vectors. A factor in the row space of the factor matrix is constant on the feasible set, and substituting it leaves a linear function. A component with three or more hyperedges is outside the hypothesis of \cref{thm:lp}. In the latent laws, each free latent factor contributes one simplex to every product term that the factor equations do not reduce, so the degree of that program is the number of free factors.
\end{proof}

\subsection{Witness expansions}
\label{app:witnesses}

\Cref{tab:casemap} lists one instance for each way the conditions of
\cref{thm:lp} can hold or fail. The paragraphs below give the expansions.

\begin{table}[ht]
\centering
\caption{Case map of \cref{thm:lp} and \cref{cor:solver}. $\pdeg$ is the
number of free latent factors after pinning and after dropping unread
$c$-components.}
\label{tab:casemap}
\small
\begin{tabular}{@{}llll@{}}
\toprule
Setting & Reason & $\pdeg$ & Program \\
\midrule
Bow or one oval, $P(Y_x)$, MTR & one $c$-component, $\lambda=1$ & $1$ & LP \\
Classical IV, $P(Y_x)$ or PNS & $\{Z\}$ not read & $1$ & LP \\
Mediated $X\leftrightarrow W\to Y$, $P(Y_x)$ & $Y$ pinned & $1$ & LP \\
Same graph, PNS & one event reads two & $2$ & bilinear \\
Pair chain, linear $Q$ & $\lambda=2$, root $X$ & $2$ & LP \\
IV with $Z\leftrightarrow X$, $P(Y_x)$ & $\lambda=2$, root $Z$ & $2$ & LP (Balke--Pearl) \\
Descendant (\cref{ex:desc}) & $\lambda=2$, no root block & $2$ & bilinear \\
Markovian, $P(Y_x)$ & all pinned & $0$ & point \\
Markovian, PNS & reads $\{Y\}$ only & $1$ & LP \\
Bow, anchored MSM & linear inequalities & $1$ & LP \\
Bow, unanchored $\Gamma$ & bilinear inequalities & $1$ & bilinear \\
Markovian chain, PNS & one event reads three & $3$ & polynomial \\
Two bows, $P(Y_x)+P(W_z)$ & one $c$-component per event & $2$ & LP (stacked) \\
Same graph, $P(Y_x,W_z)$ & one event reads two & $2$ & bilinear \\
\bottomrule
\end{tabular}
\end{table}

\paragraph{Bow / order-only.}
$X=f_X(U)$, $Y=f_Y(X,U)$. Then
$Q=\sum_u\ind[f_Y(x,u)=1]P(U{=}u)$ and
$P(x',y')=\sum_u\ind[f_X(u)=x']\ind[f_Y(x',u)=y']P(U{=}u)$.
Both are linear in one simplex. Faces (MTR, exclusions) stay linear.

\paragraph{Mediated one latent, interventional.}
This is \cref{ex:mediated}; the expansion is~\eqref{eq:mediated}.
\Cref{lem:middle} pins $Y$ for single-world use. One $c$-component,
$\{X,W\}$, is read. This instance is outside
\citet[Proposition~2]{symbolicbounds2023general}: see \cref{sec:sachs}
for the three failed bipartitions.

\paragraph{Classical IV.}
This is \cref{ex:iv-lp}; the expansion is~\eqref{eq:iv-exp}. The instrument
$\{Z\}$ is a $c$-component that the query does not read; the query reads
$\{X,Y\}$, with $\lambda=1$. PNS on the same graph reads the same
$c$-component. Two independent instruments $Z_1,Z_2$ with $Z_1\perp Z_2$ are
two unread $c$-components; their Condition~3 fails and the program remains an
LP.

\paragraph{Long Markovian PNS.}
$X=f_X(U_X)$, $M_1=f_{M_1}(X,U_1)$, $M_2=f_{M_2}(M_1,U_2)$,
$Y=f_Y(M_2,U_Y)$, all noises independent, and
$Q=P(Y_1{=}1,Y_0{=}0)$. Each of $R_{M_1}$, $R_{M_2}$, and $R_Y$ is used in two
worlds, so \cref{lem:middle} does not apply, and the single counterfactual event reads three
$c$-components: linearity fails. The query expands as a product
$P(U_1)P(U_2)P(U_Y)$, of degree $3$.

\paragraph{Markovian PNS.}
$X=f_X(U_X)$, $Y=f_Y(X,U_Y)$, $U_X\perp U_Y$, and
$Q=P(Y_1{=}1,Y_0{=}0)$. Then
$Q=\sum_{u_Y}\ind[f_Y(1,u_Y)=1]\ind[f_Y(0,u_Y)=0]P(U_Y)$.
The query reads $\{Y\}$ only, in two worlds, so $\{Y\}$ is a read
$c$-component with $\lambda=1$. The interventional query $P(Y_x)$ pins
$Y$ (g-formula) and is identified.

\paragraph{Pair chain.}
$U_1\perp U_2$ on overlapping pairs along $X\to W\to Y$:
\begin{align*}
Q
&=
\sum_{u_1,u_2}
\ind\bigl[f_Y(f_W(x,u_1,u_2),u_2)=1\bigr]
P(U_1)P(U_2),
\\
P(x',w',y')
&=
\sum_{u_1,u_2}\ind[\,\cdots\,]\,P(U_1)P(U_2).
\end{align*}
Neither factor is pinned, and the program is bilinear in $(P(U_1),P(U_2))$.
In the joint type law it is the LP of \cref{prop:lift}(a). The same directed
skeleton with one hyperedge $\{X,W,Y\}$ is the bow/oval case above.

\paragraph{Mediated graph, PNS.}
Same SCM as the mediated interventional case, but
\[
Q
=
\sum_{u,u_Y}
\ind[f_Y(W_1(u),u_Y)=1]\,
\ind[f_Y(W_0(u),u_Y)=0]\,
P(U)P(U_Y).
\]
$Y$ is used in two worlds, so \cref{lem:middle} does not remove $U_Y$, and the
counterfactual event reads $\{X,W\}$ and $\{Y\}$. Two free factors.

\paragraph{Additive two bows.}
The expansions are in \cref{ex:two-bows}. Each counterfactual event of $Q$ reads one
$c$-component, and the constraints are the two $c$-factors, each
linear in one simplex. The stacked program
$\min/\max\,c_1^\top q_1+c_2^\top q_2$ subject to
$A_1 q_1=P(X,Y)$ and $A_{2,y}q_2=P(Z,W\mid Y{=}y)$ is an LP. The conjunction
$P(Y_x{=}1,W_z{=}1)$ on the same graph is $(c_1^\top q_1)(c_2^\top q_2)$,
bilinear in these coordinates.

\paragraph{A descendant that matters.}
This is \cref{ex:desc}. By \cref{lem:two}, with $S_1=\{Y\}$, $S_{12}=\{X\}$,
$S_2=\{D\}$, the model is the set of laws $q(r_Y,x,r_D)$ with
$\sum_x q(r_Y,x,r_D)=a(r_Y)\,\nu(r_D)$, where $a$ and $\nu$ are the laws of
$R_Y$ and $R_D$, and
\begin{align*}
P(Y_1{=}1)&=\sum_{r_Y,x,r_D}\ind[r_Y(1)=1]\,q(r_Y,x,r_D),\\
P(x,y,d)&=\sum_{r_Y,r_D}\ind[r_Y(x)=y]\,\ind[r_D(y)=d]\,q(r_Y,x,r_D).
\end{align*}
For a fixed $\nu$ the program is an LP in $(q,a)$. Optimizing that LP over a
grid of $\nu$ gives witness models, hence values inside the identified
interval. A branch and bound over boxes of $\nu$, with the McCormick envelopes
of the products $a(r_Y)\nu(r_D)$ on each box (an LP relaxation), gives valid
outer values. Run until the two agree to $5\cdot10^{-4}$, they place the lower
bound in $[0.0213,0.0218]$ and the upper bound in $[0.5818,0.5823]$, against
$[0.017,0.618]$ after dropping $D$.

\subsection{Anchored MSM and unanchored Rosenbaum bounds}
\label{app:msm}

Write $\pi_0(\,\cdot\mid\mathrm{pa})$ for the observed propensity.
The anchored Tan bound is
\[
\Lambda^{-1}
\le
\frac{\pi(x\mid r,\mathrm{pa})/\pi(x'\mid r,\mathrm{pa})}
{\pi_0(x\mid\mathrm{pa})/\pi_0(x'\mid\mathrm{pa})}
\le
\Lambda
\]
whenever the observed odds are defined. Multiplying through by the
(constant) observed odds and by the type masses gives
\[
q(x,r;\mathrm{pa})\,\pi_0(x'\mid\mathrm{pa})
\le
\Lambda\,\pi_0(x\mid\mathrm{pa})\,q(x',r;\mathrm{pa}),
\]
linear in $q$. For binary root $X$ this is
$q(1,r)\le\Lambda\theta_0 q(0,r)$ and
$q(0,r)\le(\Lambda/\theta_0)q(1,r)$ with
$\theta_0=e_0/(1-e_0)$. The $\ell_\infty$ $c$-dependence bound
$\lvert e(r)-e_0\rvert\le c$ likewise clears to linear inequalities.

The unanchored Rosenbaum comparison~\eqref{eq:gamma} is the same odds
bound without a data-constant baseline: the product of two free
coordinates remains. Anchoring to the observed odds within a factor
$\sqrt{\Gamma}$ is Tan MSM at $\Lambda=\sqrt{\Gamma}$.

On~\eqref{eq:re1} one has $e_0=1/2$ and $\theta_0=1$. Writing
$S=q(0,01)+q(0,11)$ and $Q=p_{11}+S$, the MSM couplings give
\begin{align*}
S_{\max}(\Lambda)
&=
\min\bigl\{P(X{=}0),\,\Lambda p_{11},\,P(X{=}0)-p_{10}/\Lambda\bigr\},
\\
S_{\min}(\Lambda)
&=
\max\bigl\{0,\,p_{11}/\Lambda,\,P(X{=}0)-\Lambda p_{10}\bigr\}.
\end{align*}
Substituting the cells of~\eqref{eq:re1} produces
\[
u(\Lambda)=0.85-\frac{0.15}{\Lambda},
\qquad
\ell(\Lambda)
=
\begin{cases}
0.85-0.15\Lambda, & 1\le\Lambda\le 7/3,\\
0.35\,(1+\Lambda^{-1}), & \Lambda\ge 7/3.
\end{cases}
\]

\begin{figure}[t]
\centering
\begin{tikzpicture}[x=1.45cm,y=6.0cm,>=Stealth]
  \fill[blue!8] plot[domain=1:7/3,samples=30] (\x,{0.85-0.15/\x})
     -- plot[domain=7/3:1,samples=30] (\x,{0.85-0.15*\x}) -- cycle;
  \fill[blue!8] plot[domain=7/3:7,samples=40] (\x,{0.85-0.15/\x})
     -- plot[domain=7:7/3,samples=40] (\x,{0.35*(1+1/\x)}) -- cycle;
  \draw[->] (0.8,0.30) -- (7.4,0.30) node[right,font=\small] {$\Lambda$};
  \draw[->] (0.8,0.30) -- (0.8,0.92) node[above,font=\small] {$P(Y_1{=}1)$};
  \foreach \x in {1,2,3,4,5,6,7}
    \draw (\x,0.30) -- (\x,0.292) node[below,font=\scriptsize] {\x};
  \foreach \y in {0.35,0.40,0.55,0.70,0.85}
    \draw (0.8,\y) -- (0.77,\y) node[left,font=\scriptsize] {\y};
  \draw[dashed,black!40] (0.8,0.85) -- (7.2,0.85);
  \draw[dashed,black!40] (0.8,0.35) -- (7.2,0.35);
  \draw[dotted,black!60] (0.8,0.40) -- (7.2,0.40);
  \draw[very thick,blue!70!black,domain=1:7,samples=60] plot (\x,{0.85-0.15/\x});
  \draw[very thick,red!70!black,domain=1:7/3,samples=30] plot (\x,{0.85-0.15*\x});
  \draw[very thick,red!70!black,domain=7/3:7,samples=40] plot (\x,{0.35*(1+1/\x)});
  \fill (1,0.70) circle (1.5pt);
  \node[font=\scriptsize,anchor=south west] at (5.0,0.825) {$u(\Lambda)$};
  \node[font=\scriptsize,anchor=north west] at (3.6,0.455) {$\ell(\Lambda)$};
  \node[font=\scriptsize,anchor=west,black!60] at (7.2,0.85) {Manski $u$};
  \node[font=\scriptsize,anchor=west,black!60] at (7.2,0.35) {Manski $\ell$};
  \node[font=\scriptsize,anchor=west,black!60] at (7.2,0.40) {untreated rate};
\end{tikzpicture}
\caption{Sharp interval for $P(Y_1{=}1)$ on~\eqref{eq:re1} under an anchored
Tan budget $\Lambda$. At $\Lambda=1$ the effect is identified at $0.70$; the
lower bound falls to the untreated employment rate $0.40$ at $\Lambda=7$.}
\label{fig:msm-intro}
\end{figure}
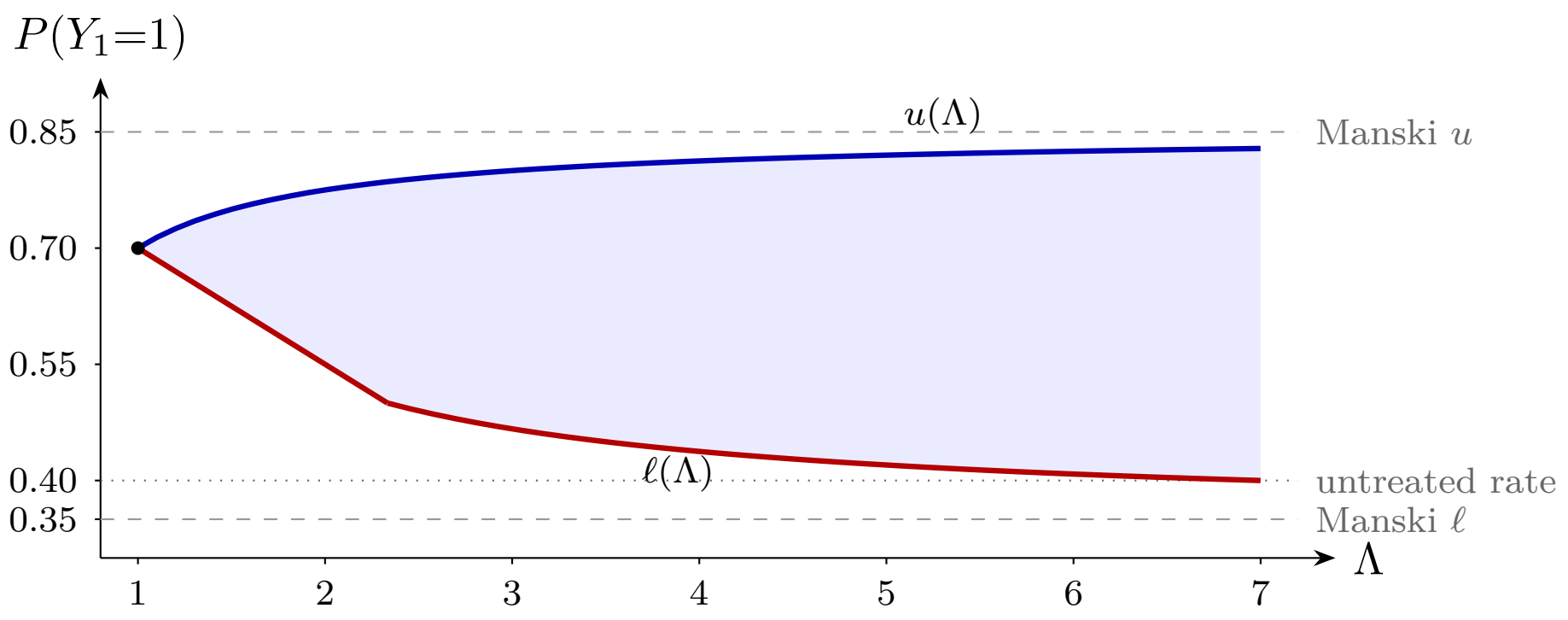

\subsection{Manski and MTR}
\label{app:manski}

On a binary bow the natural bounds and the monotone-treatment-response bounds are
\begin{equation}
\label{eq:manski}
P(X{=}x,Y{=}y)\;\le\;P(Y_x{=}y)\;\le\;P(X{=}x,Y{=}y)+1-P(X{=}x),
\end{equation}
\begin{equation}
\label{eq:mtr}
P(Y{=}1)\;\le\;P(Y_1{=}1)\;\le\;P(X{=}1,Y{=}1)+P(X{=}0).
\end{equation}

\begin{proof}[Proof of~\eqref{eq:manski} and~\eqref{eq:mtr}]
On the binary bow the observational map partitions types into four
fibers, one per cell of $P$. For $Q=P(Y_1{=}1)$, types in the
$X{=}1$ fibers have $Y_1$ pinned to the observed $Y$; types in the
$X{=}0$ fibers have $Y_1$ free in $\{0,1\}$. Maximizing $Q$ assigns
$Y_1{=}1$ on all free mass $P(X{=}0)$; minimizing assigns $Y_1{=}0$.
Hence $\ell=p_{11}$, $u=p_{11}+P(X{=}0)$, width $1-P(X{=}1)$. For finite
$X$ and $Y$ the same fiber argument applies: types with $X=x$ have $Y_x$
pinned to the observed $Y$, and types with $X\neq x$ leave $Y_x$ free. This
gives $P(x,y)\le P(Y_x{=}y)\le P(x,y)+1-P(X{=}x)$.

MTR zeros every type with $Y_1<Y_0$. A unit in the cell
$(X,Y)=(0,1)$ already has $Y_0{=}1$, so $Y_1$ is forced to $1$ and
that cell is no longer free. The only remaining free mass is
$P(X{=}0,Y{=}0)$, and the width is $p_{00}$. On~\eqref{eq:re1} this
is the interval $[0.55,0.85]$.
\end{proof}

\subsection{Monotonicity and interpolation}
\label{app:mono}

Intersecting the feasible set with a further constraint can only shrink
the image of $Q$, or make the program infeasible.

\begin{corollary}[Interpolation of linear or convex budgets]
\label{cor:interp}
For an anchored linear or convex sensitivity model with budget
$\rho$, the width $w(P;a,\rho)$ is nondecreasing in $\rho$. It
interpolates the no-unobserved-confounding pole (when that pole is
point-identified for an interventional query) and the Manski width as
$\rho\to\infty$.
\end{corollary}

\begin{proof}
Nested feasible sets give monotonicity: a larger budget admits every
model the smaller budget admits.
Continuity in $\rho$ for linear faces follows because the
right-hand side depends continuously (in fact linearly) on the
budget. The two poles are the Markovian program ($\rho$ at its
no-confounding value) and the order-only program ($\rho\to\infty$).
\end{proof}

\clearpage
\section{Further instances of the theorem}
\label{sec:instances}

The expansions below are the checks that the conditions of \cref{thm:lp}
are met, or fail, on the graphs used in the text.

\subsection{Independent latents}
\label{sec:product}

Confounding is recorded by \emph{latent hyperedges}: each hyperedge $h$ is one
exogenous source that enters the equations of every variable in $h$. A
bidirected edge $A\leftrightarrow B$ is the pair $\{A,B\}$. Two overlapping
bidirected edges are two hyperedges, not one. The connected components of the
undirected graph with those edges as links are the
\emph{$c$-components} \citep{tian2002general}. Distinct hyperedges have
independent latents by default. The canonical law is the tensor product of one
probability vector per hyperedge (and one for the private noise of each
unconfounded variable). We call this the \emph{hyperedge product
parameterization}. It is not the same as placing one free joint type law on
each $c$-component, as in \cref{def:canonical_scm}: a pair-chain
$c$-component with two hyperedges is a product of two simplices, whereas a
single oval over the same nodes is one simplex.

For a $c$-component $C$ write $\lambda(C)$ for the number of independent
latents that enter it: the number of hyperedges that meet $C$, or $1$ if
$C=\{V\}$ is an unconfounded singleton (its private noise). A bow, an oval, and
an unconfounded singleton have $\lambda=1$; the pair chain has $\lambda=2$.

\begin{figure}[t]
\centering
\begin{tikzpicture}[>=Stealth, every node/.style={circle,draw,inner sep=1.5pt,font=\small,minimum size=1.15em}]
  \begin{scope}[local bounding box=A]
    \node (X) at (0,0) {$X$};
    \node (Y) at (1.6,0) {$Y$};
    \draw[->] (X) -- (Y);
    \draw[<->,dashed] (X) to[bend left=42] (Y);
  \end{scope}
  \node[font=\small,anchor=north,draw=none] at ($(A.south)+(0,-0.15)$) {(a) bow};

  \begin{scope}[shift={(4.0,0)}, local bounding box=B]
    \node (Z) at (0,0) {$Z$};
    \node (X) at (1.5,0) {$X$};
    \node (Y) at (3.0,0) {$Y$};
    \draw[->] (Z) -- (X);
    \draw[->] (X) -- (Y);
    \draw[<->,dashed] (X) to[bend left=42] (Y);
  \end{scope}
  \node[font=\small,anchor=north,draw=none] at ($(B.south)+(0,-0.15)$) {(b) IV};

  \begin{scope}[shift={(0,-2.35)}, local bounding box=C]
    \node (X) at (0,0) {$X$};
    \node (W) at (1.5,0) {$W$};
    \node (Y) at (3.0,0) {$Y$};
    \draw[->] (X) -- (W);
    \draw[->] (W) -- (Y);
    \draw[<->,dashed] (X) to[bend left=40] (W);
    \draw[<->,dashed] (W) to[bend left=40] (Y);
  \end{scope}
  \node[font=\small,anchor=north,draw=none] at ($(C.south)+(0,-0.15)$) {(c) pair chain};

  \begin{scope}[shift={(5.2,-2.35)}, local bounding box=D]
    \node (X) at (0,0) {$X$};
    \node (W) at (1.5,0) {$W$};
    \node (Y) at (3.0,0) {$Y$};
    \draw[->] (X) -- (W);
    \draw[->] (W) -- (Y);
    \draw[dashed] (0.15,0.55) .. controls (1.5,1.25) .. (2.85,0.55);
    \node[draw=none,font=\scriptsize,black!55] at (1.5,1.22) {$U$};
  \end{scope}
  \node[font=\small,anchor=north,draw=none] at ($(D.south)+(0,-0.15)$) {(d) one oval};
\end{tikzpicture}
\caption{Four latent drawings. (a)~One shared latent: LP in a single simplex.
(b)~Unconfounded instrument and a bow on $\{X,Y\}$: two $c$-components, of
which the query reads only $\{X,Y\}$. (c)~Two pairwise latents: a product of
two simplices in the latent laws, yet still an LP in the joint law of the
response types (\cref{prop:lift}). (d)~Same directed chain, one shared latent:
one simplex, LP. The directed skeleton does not decide the number of
factors.}
\label{fig:four-graphs}
\end{figure}
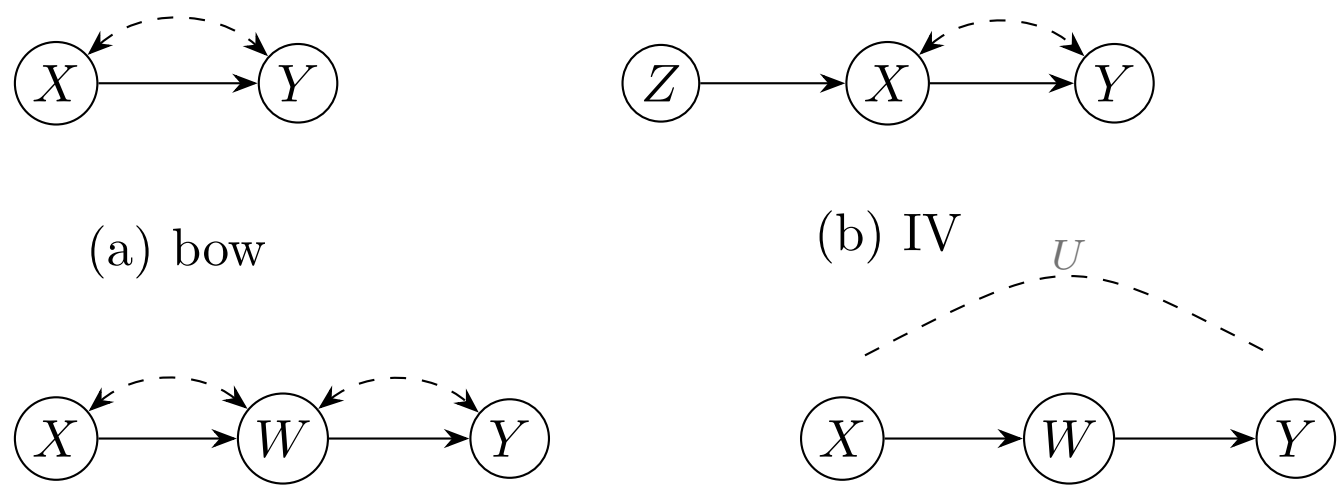

\begin{table}[ht]
\centering
\caption{Response types of the binary bow $\G:\{X\to Y,\, X\leftrightarrow Y\}$.
$R_X$ is the value of $X$, and $R_Y=(Y_0,Y_1)$. The factual cell is
$(X,Y)=(R_X,Y_{R_X})$.}
\label{tab:types}
\small
\begin{tabular}{@{}cccccc@{}}
\toprule
$(R_X,R_Y)$ & $X$ & $Y_0$ & $Y_1$ & Factual cell & In $P(Y_1{=}1)$? \\
\midrule
$(0,00)$ & $0$ & $0$ & $0$ & $(0,0)$ & no \\
$(0,01)$ & $0$ & $0$ & $1$ & $(0,0)$ & yes \\
$(0,10)$ & $0$ & $1$ & $0$ & $(0,1)$ & no \\
$(0,11)$ & $0$ & $1$ & $1$ & $(0,1)$ & yes \\
$(1,00)$ & $1$ & $0$ & $0$ & $(1,0)$ & no \\
$(1,01)$ & $1$ & $0$ & $1$ & $(1,1)$ & yes \\
$(1,10)$ & $1$ & $1$ & $0$ & $(1,0)$ & no \\
$(1,11)$ & $1$ & $1$ & $1$ & $(1,1)$ & yes \\
\bottomrule
\end{tabular}
\end{table}

\begin{example}[Encouragement design: reduction to Balke--Pearl]
\label{ex:iv-lp}
A randomized offer $Z$ encourages enrollment $X$ in a programme with outcome
$Y$ (\cref{sec:ill-designs}). This is the left/right split of
\citet[Proposition~2]{symbolicbounds2023general}.
The equations are $Z=f_Z(U_Z)$, $X=f_X(Z,U)$, $Y=f_Y(X,U)$ with
$U_Z\perp U$ (\cref{fig:four-graphs}b). The $c$-components are $\{Z\}$
(their $W_L$) and $\{X,Y\}$ (their $W_R$): every arrow between them is
$Z\to X$, and no latent meets both. The
instrument has its own type $R_Z\in\{0,1\}$, pinned by $P(Z)$. The
types of $\{X,Y\}$ are the eight bow types of \cref{tab:types}, except
$R_X$ is now a map $\{0,1\}\to\{0,1\}$ (four compliance types:
never-taker, complier, defier, always-taker), so this $c$-component has
$4\times 4=16$ masses $q(r_X,r_Y)$. Expand:
\begin{align}
\label{eq:iv-exp}
P(Y_x{=}y)
&=
\sum_{r_X,r_Y}
\ind[r_Y(x)=y]\,q(r_X,r_Y),
\\
P(x,y\mid z)
&=
\sum_{r_X,r_Y}
\ind[r_X(z)=x]\,\ind[r_Y(x)=y]\,q(r_X,r_Y).
\nonumber
\end{align}
The query does not depend on $R_Z$. For each level $z$ the conditional
$P(x,y\mid z)$ is a linear image $A_z q$ of the same $q$, exactly as
their $p=Pq$. The joint factors as
$P(z,x,y)=P(z)\,P(x,y\mid z)$; matching the joint is equivalent to
matching those conditionals. The reduced program is therefore the LP
\begin{equation}
\label{eq:iv-lp}
\min/\max_q\;
c^\top q
\quad\text{s.t.}\quad
A_z q=P(\,\cdot\mid z)
\text{ for each }z,\quad
q\ge 0,\quad
\mathbf{1}^\top q=1.
\end{equation}
This is Balke--Pearl. The same reduction holds when the instrument block is
itself confounded (their Condition~3 fails): it is then a $c$-component with
its own latent that the query does not read, and it only has to reproduce
$P(z)$ (\cref{thm:lp}).
\end{example}

\begin{example}[Pair chain]
\label{ex:pair}
On $X\to W\to Y$ with hyperedges $\{X,W\}$ and $\{W,Y\}$
(\cref{fig:four-graphs}c) the two latents $U_1\perp U_2$ both meet $W$, so
$\{X,W,Y\}$ is one $c$-component with $\lambda=2$. Evaluating along the
chain,
\begin{align}
\label{eq:pair}
P(x,w,y)
&=
\sum_{u_1,u_2}
\ind[f_X(u_1)=x]\,
\ind[f_W(x,u_1,u_2)=w]\,
\ind[f_Y(w,u_2)=y]\,
p_1(u_1)p_2(u_2),
\\
P(Y_x{=}1)
&=
\sum_{u_1,u_2}
\ind\bigl[f_Y\bigl(f_W(x,u_1,u_2),u_2\bigr)=1\bigr]
p_1(u_1)p_2(u_2).
\nonumber
\end{align}
Every summand is a product $p_1(u_1)p_2(u_2)$, and neither factor is
pinned by a margin of $P$: written in $(p_1,p_2)$ the program is bilinear.
In the joint law of the response types the same constraint is the linear
independence of \cref{lem:two}, so the sharp program is an LP
(\cref{prop:lift}). The same directed chain with one hyperedge
$\{X,W,Y\}$ (\cref{fig:four-graphs}d) is one simplex.
\end{example}

\subsection{Query and assumption classes}
\label{sec:labels}

Solver choice depends on the graph and on the algebraic class of the query
and of the assumptions. Linear assumptions are \cref{def:labels}, and
syntactically linear queries (\cref{def:linear-query}) are distinct from
linearity of the objective after pinning and removal (\cref{def:linearity}).

A counterfactual event is a conjunction of factual or potential-outcome literals
(nested potential outcomes are allowed as syntax). Linear queries (\cref{def:linear-query})
therefore include single-world effects $P(Y_x)$ and
$\mathrm{ATE}=P(Y_1)-P(Y_0)$; cross-world events such as
$\mathrm{PNS}=P(Y_1{=}1,Y_0{=}0)$; PN and PS, whose denominator is an
observational cell of $P$; and sums of counterfactual events such as $P(Y_x)+P(W_z)$.
Outside the class are a product of queries, a nonlinear functional of
several counterfactual events, and a conditional whose denominator is a free combination
of type masses.

\begin{example}[Unanchored Rosenbaum $\Gamma$]
\label{ex:gamma}
Unanchored Rosenbaum $\Gamma$ is the witness that one simplex is not enough:
on the single simplex of \cref{ex:bow} the pairwise odds bound
\begin{equation}
\label{eq:gamma}
q(1,r)\,q(0,r')
\le
\Gamma\,q(0,r)\,q(1,r')
\end{equation}
is bilinear in $q$. Anchoring every type's odds to the observed propensity
removes the product and yields the Tan MSM of \cref{fig:msm-intro}
(\cref{app:msm}).
\end{example}

\begin{example}[Two bows: additive versus conjunctive queries]
\label{ex:two-bows}
Take $X=f_X(U_1)$, $Y=f_Y(X,U_1)$, $Z=f_Z(Y,U_2)$, $W=f_W(Z,U_2)$ with
$U_1\perp U_2$: two bows $\{X,Y\}$ and $\{Z,W\}$ joined by $Y\to Z$.
For $Q=P(Y_x{=}1)+P(W_z{=}1)$,
\begin{align*}
P(Y_x{=}1)
&=
\sum_{u_1}\ind[f_Y(x,u_1)=1]\,P(U_1{=}u_1),
&
P(W_z{=}1)
&=
\sum_{u_2}\ind[f_W(z,u_2)=1]\,P(U_2{=}u_2),
\end{align*}
so $Q=c_1^\top q_1+c_2^\top q_2$ and each counterfactual event reads one $c$-component. The
observational law factors as $P(x,y,z,w)=P(x,y)\,P(z,w\mid y)$ with
\begin{align*}
P(x,y)&=\sum_{u_1}\ind[f_X(u_1)=x]\,\ind[f_Y(x,u_1)=y]\,P(U_1{=}u_1),\\
P(z,w\mid y)&=\sum_{u_2}\ind[f_Z(y,u_2)=z]\,\ind[f_W(z,u_2)=w]\,P(U_2{=}u_2),
\end{align*}
one $c$-factor per bow, each linear in one simplex. The program is two
Balke--Pearl LPs stacked side by side.
The \emph{conjunction} $P(Y_x{=}1,W_z{=}1)=(c_1^\top q_1)(c_2^\top q_2)$
on the same graph is one counterfactual event that reads both $c$-components.
The equations of either bow do not fix the corresponding factor, so linearity
fails and \cref{thm:lp} does not apply.
\end{example}

Syntactic linearity (\cref{def:linear-query}) is not a licence for every counterfactual. Mediated PNS
on $X\leftrightarrow W\to Y$ is one counterfactual event $P(Y_1{=}1,Y_0{=}0)$ that reads
$\{X,W\}$ and evaluates $Y$ in two worlds, so pinning fails on $Y$ and linearity (\cref{def:linearity}) fails.

\paragraph{Two coupled hyperedges.}
\Cref{lem:two} turns two instances that a product parameterization would
call bilinear into LPs.

\begin{corollary}[Exact LPs for two coupled hyperedges]
\label{prop:lift}
\leavevmode
\begin{enumerate}[label=(\alph*),leftmargin=1.6em,itemsep=0.2em]
\item \emph{Pair chain} (\cref{ex:pair}). Here $S_1=\{X\}$, $S_{12}=\{W\}$,
$S_2=\{Y\}$, and $X$ is a root. For every linear query on $(X,W,Y)$---in
particular $P(Y_x)$, the ATE, and PNS---the sharp bounds are the LP of
\cref{thm:lp} in the masses $q(x,r_W,r_Y)$, with the single constraint
\begin{equation}
\label{eq:lift}
\sum_{r_W}q(x,r_W,r_Y)=P(X{=}x)\sum_{x',r_W}q(x',r_W,r_Y)
\quad\text{for all }x,r_Y.
\end{equation}
\item \emph{Instrument confounded with the treatment}: $Z\to X\to Y$ with
hyperedges $\{Z,X\}$ and $\{X,Y\}$, and no $Z\to Y$. Here $S_1=\{Z\}$ is a
root, so the sharp bounds are an LP. For every query that depends only on
the outcome's response type---$P(Y_x)$, the ATE, PNS---they equal the
Balke--Pearl LP~\eqref{eq:iv-lp} of \cref{ex:iv-lp}.
\end{enumerate}
\end{corollary}

\begin{proof}[Proof of \cref{prop:lift}]
\emph{(a) Pair chain.}
\textit{SCM.} $X=f_X(U_1)$, $W=f_W(X,U_1,U_2)$, $Y=f_Y(W,U_2)$, with
$U_1\perp U_2$; $S_1=\{X\}$, $S_{12}=\{W\}$, $S_2=\{Y\}$.
\textit{Query and data.} In the masses $q(x,r_W,r_Y)$,
\begin{align*}
P(Y_x{=}y)&=\sum_{x',r_W,r_Y}\ind[r_Y(r_W(x))=y]\,q(x',r_W,r_Y),\\
P(x,w,y)&=\sum_{r_W,r_Y}\ind[r_W(x)=w]\,\ind[r_Y(w)=y]\,q(x,r_W,r_Y),
\end{align*}
and PNS is the same sum with $\ind[r_Y(r_W(1))=1,\,r_Y(r_W(0))=0]$.
\textit{Reduction.} $X$ is a root, so \cref{lem:two} gives the linear
constraint~\eqref{eq:two}, which here is~\eqref{eq:lift}. The query reads one
$c$-component, and \cref{thm:lp} applies.

\emph{(b) Instrument confounded with the treatment.}
\textit{SCM.} $Z=f_Z(U_1)$, $X=f_X(Z,U_1,U_2)$, $Y=f_Y(X,U_2)$, with
$U_1\perp U_2$; $S_1=\{Z\}$ is a root, so \cref{thm:lp} applies as in~(a).

\textit{Comparison with Balke--Pearl.} The classical IV model of
\cref{ex:iv-lp} imposes $Z\perp(R_X,R_Y)$, which is stronger than
$Z\perp R_Y$, so its interval is contained in ours. Conversely, take a
feasible $q(z,r_X,r_Y)$, write $\nu$ for the law of $R_Y$, and set
$\kappa_z(x\mid r_Y)=q(X{=}x\mid Z{=}z,R_Y{=}r_Y)$. Put
$\tilde q(r_X,r_Y)=\nu(r_Y)\prod_z\kappa_z(r_X(z)\mid r_Y)$, which draws the
coordinates $r_X(z)$ independently given $R_Y$. Then
$\tilde q(r_X(z){=}x,\,r_Y)=\nu(r_Y)\,\kappa_z(x\mid r_Y)=q(X{=}x,r_Y\mid z)$,
using $Z\perp R_Y$. So $\tilde q$ reproduces every $P(x,y\mid z)$ with the same law
of $R_Y$. The two models produce the same pairs of data and laws of $R_Y$, and
any query that depends only on $R_Y$ has the same sharp bounds in both.
\end{proof}

Part~(b) says that letting the instrument share a latent with the
treatment (but not with the outcome) leaves the Balke--Pearl bounds
unchanged; on the instrument table of \cref{sec:ill-designs} the package
returns the Balke--Pearl interval $[0.283,0.615]$ for both graphs. For
part~(a), the table is~\eqref{eq:re2}, whose $(X,Y)$-margin is~\eqref{eq:re1}. The LP of part~(a)
gives $[0.350,0.850]$ for $P(Y_1{=}1)$ and $[0.000,0.600]$ for PNS. One
hyperedge on $\{X,W,Y\}$ gives the same interval for $P(Y_1{=}1)$ but
$[0.000,0.650]$ for PNS: here the independence of the two latents narrows
PNS but not the interventional probability.

\paragraph{Non-ancestors.}
The query only uses the mechanisms of $\An(V_Q)$, yet the data of other
variables can matter.

\begin{proposition}[Dropping non-ancestors]
\label{prop:trim}
Let $A=\An(V_Q)$. The interval computed from the subgraph induced by $A$
(each hyperedge intersected with $A$) and the margin $P(\mathbf{v}_A)$
contains the identified interval. If no hyperedge meets both $A$ and its
complement and the full program is feasible, the two intervals coincide.
\end{proposition}

\begin{proof}[Proof of \cref{prop:trim}]
Take any model on $\G$ that reproduces $P$ and obeys the assumptions. The variables of
$A=\An(V_Q)$ read only parents in $A$ and the latents of hyperedges that meet
$A$. Keep those mechanisms and latents, and intersect each hyperedge with
$A$. The result is a model on the induced subgraph that reproduces
$P(\mathbf{v}_A)$ and has the same value of $Q$, because $Q$ uses only the
mechanisms of $A$. So every value in the identified interval is feasible for
the trimmed program. If no hyperedge meets both $A$ and its complement, $A$ is
a union of $c$-components, and by Step~1 of the proof of \cref{thm:lp} the
complement is a separate block that the query does not read: any trimmed model
combines with a feasible complement block into a full model.
\end{proof}

\begin{example}[A descendant that matters]
\label{ex:desc}
Take $X\to Y\to D$ with hyperedges $\{X,Y\}$ and $\{X,D\}$, and
$Q=P(Y_1{=}1)$. The equations are $X=f_X(U_1,U_2)$, $Y=f_Y(X,U_1)$,
$D=f_D(Y,U_2)$ with $U_1\perp U_2$. The query
$P(Y_1{=}1)=\sum_{u_1}\ind[f_Y(1,u_1)=1]P(U_1{=}u_1)$ does not involve $U_2$,
and $D$ is not an ancestor of $Y$. Dropping $D$ leaves the bow, whose sharp
interval is Manski on $P(x,y)$. But $\{X,Y,D\}$ is one $c$-component with
$\lambda=2$, $S_1=\{Y\}$, $S_{12}=\{X\}$, $S_2=\{D\}$: the treatment reads
$U_2$, the descendant $D$ reveals information about $U_2$, and the
independence $R_Y\perp R_D$ of \cref{lem:two} then restricts how $X$ can
depend on $R_Y$. Neither hyperedge contains both variables with a parent, so covering fails and the
constraint stays bilinear. On the table with cells
$P(x,y,d)$, in the order $000,001,\ldots,111$,
$(0.046,0.010,0.047,0.498,0.103,0.279,0.010,0.007)$, dropping $D$ gives
$[0.017,0.618]$, whereas the full model gives an interval inside
$[0.021,0.583]$ (\cref{app:witnesses}).
\end{example}

\paragraph{Outside the theorem.}
When a component has three or more hyperedges, \cref{thm:lp} does not apply;
\cref{app:beyond} records the obstructions that survive at every size.
Within two hyperedges, failure of linearity leaves a product of the remaining
response laws. \Cref{prop:lift} is the covering case of two latents: the
program is linear in the joint response types, while a product parameterization
of the same latents is bilinear.

\subsection{Beyond two hyperedges}
\label{app:beyond}

The three properties used in the proof of \cref{thm:lp} --- a saturated removed component, a polytope fiber for every table, and a linear objective --- remain necessary and sufficient on any diagram. The graphical tests are proved when every component has at most two hyperedges. Past that bound, two obstructions to saturation survive at every number of hyperedges.

If $A$ and $B$ lie in one component, no hyperedge contains both, and neither $A\to B$ nor $B\to A$ is present, the table of \cref{lem:cross} is impossible. The same table, with an outside parent binary and equal to $1$ with probability $\tfrac12$ and with the downstream variable set equal to it, shows that reachability stays necessary at every number of hyperedges.

If $S$ is a set of two or more variables in the component, no one an ancestor of another, and no hyperedge contains $S$, the table in which every variable of $S$ equals one binary variable that equals $1$ with probability $\tfrac12$, and every other variable equals $0$, is impossible. A parent of a variable in $S$ cannot lie in $S$, so every such parent equals $0$, and each variable of $S$ is a function of the hidden causes through it. Those functions agree. \Cref{lem:agree}, applied to each pair and then through $S$, puts that common value on a hidden cause whose hyperedge contains every member of $S$. There is no such hyperedge.

Both obstructions are necessary. They are not sufficient at four variables. The chain $A\to B\to C\to D$ with hyperedges $\{A,C\}$, $\{A,D\}$, and $\{B,D\}$ has a direct edge or a shared hyperedge for every pair, and a chain has no antichain of size two or more. It is not saturated. The equations are $A=f(U,V)$, $B=\beta(W)$, $C=h(B,U)$, and $D=k(C,V,W)$, with $U\perp V\perp W$. If $B\perp A$ and $D=A$, then at each value of $B$ the variable $A$ is a function of $(h(B,U),V)$, so \cref{lem:gap} applies with conditioning variable $B$ and inner variable $C$. The absolute gap of $P(C\mid A,B)$ is the same at every value of $B$, or both gaps are $0$. The binary table with $A$ and $B$ independent and each equal to $1$ with probability $\tfrac12$, $D=A$, and gaps $-0.7$ and $-0.5$ is therefore impossible.

On the fiber side, the same boundary is open. If at least two variables have a parent and every hyperedge misses at least one of them, the zero configuration has a fiber that is not a polytope: the point mass on the constant-$0$ response and the point mass on the response that sends one parent value $0$ to $0$ and another to $1$ both reproduce that table, and their average shares one bit across every variable with a parent. \Cref{lem:agree} puts that bit on a cause heard by all of them, and no hyperedge contains them all. A graphical list of the components whose fiber is a polytope for every table, once three or more hyperedges are allowed, is open.

\subsection[Relation to Sachs et al.]{Relation to \citet{symbolicbounds2023general}}
\label{sec:sachs}

\citet[Proposition~2]{symbolicbounds2023general} give a sufficient class for
a tight linear program in a joint type law $q=P(R_R)$. They require a
bipartition $\mathbf{W}=W_L\cup W_R$ such that
\begin{enumerate}[label=(\arabic*),leftmargin=1.6em,itemsep=0.15em]
\item every edge joining $W_L$ to $W_R$ is directed $W_L\to W_R$;
\item no unmeasured variable has children in both sides;
\item all of $W_L$ share one unmeasured parent $U_L$;
\item all of $W_R$ share one unmeasured parent $U_R$.
\end{enumerate}
The query must then be a linear functional of that same $q$.
Conditions~(1)--(2) say that $W_L$ is a union of $c$-components that the
query does not read. Conditions~(3)--(4) say that each side carries one
latent. \Cref{thm:lp} drops both: the unread block need not be one latent,
further $c$-components may be pinned (\cref{lem:middle}), a read
$c$-component may carry two latents (\cref{lem:two}), and identification may
reduce the program to a point. Every instance that meets their four
conditions with a linear-in-$q_R$ query meets \cref{thm:lp}. The inclusion
is strict.

\begin{example}[Mediated confounding: linear here, outside their class]
\label{ex:mediated}
Take $X\to W\to Y$ with a hyperedge only on $\{X,W\}$ (the directed chain of
\cref{fig:four-graphs}c with the $\{W,Y\}$ edge deleted). The equations are
$X=f_X(U)$, $W=f_W(X,U)$, $Y=f_Y(W,U_Y)$ with $U\perp U_Y$. The query is
$Q=P(Y_x{=}1)$. Expanding,
\begin{align}
\label{eq:mediated}
P(Y_x{=}1)
&=
\sum_w P(Y{=}1\mid W{=}w)\,P(W_x{=}w),
\\
P(W_x{=}w)
&=
\sum_u
\ind[f_W(x,u)=w]\,P(U{=}u).
\nonumber
\end{align}
The outcome is used in one world, so \cref{lem:middle} pins it to the
kernel $P(Y\mid W)$, and the model requires $Y\perp X\mid W$. What remains
is an LP in the single simplex $P(U)$: a Balke--Pearl program on the
$\{X,W\}$ bow, averaged against the observed kernel. The query reads one
$c$-component besides the pinned $Y$, so \cref{thm:lp} applies.

No bipartition satisfies their (1)--(4) with $Q$ linear in $q_R$.
\begin{itemize}[leftmargin=1.4em,itemsep=0.15em]
\item $W_L=\emptyset$, $W_R=\{X,W,Y\}$. Condition~(4) fails: $Y$ does not
share $U$ with $X$ and $W$.
\item $W_L=\{X\}$, $W_R=\{W,Y\}$. Condition~(2) fails: the latent on
$\{X,W\}$ meets both sides.
\item $W_L=\{X,W\}$, $W_R=\{Y\}$. Conditions~(1)--(4) can hold, but
$Q=P(Y_x)$ is not a linear functional of $q_Y=P(R_Y)$ alone: it depends on
the law of $W_x$, which lives on the $L$-side types.
\end{itemize}
Running their Algorithm~1 on a free joint over $(R_X,R_W,R_Y)$ would allow
$U_Y$ to depend on $U$---a different model, not sharp for $U\perp U_Y$. On
the table~\eqref{eq:re3} of \cref{sec:ill-solver} the free joint gives
$[0.270,0.770]$, whereas the sharp interval under $U\perp U_Y$ is
$[0.470,0.570]$.
\end{example}

A smaller gap is their Condition~(3) alone. Two independent instruments
$Z_1,Z_2\to X\to Y$ with $X\leftrightarrow Y$ and $Z_1\perp Z_2$ violate
(3) ($Z_1$ and $Z_2$ do not share one $U_L$), yet they are two
$c$-components that the query does not read, and \cref{thm:lp} still yields
the Balke--Pearl LP in $P(U)$.

\clearpage

\section{The \pid{} package}
\label{sec:package}
\label{sec:routing}

\pid{} implements \cref{thm:lp} for counterfactual queries under assumptions the user composes. The call takes four objects: finite variables, an observational table, a query, and an assumption set, and returns an interval together with a flag that says whether the interval is the identified set.

For the bow of \cref{ex:bow} under monotone treatment response the call is
\begin{lstlisting}
from partialid import Assumptions, Query, Variables, bound

V = Variables.from_binary("X", "Y")
P = {(0, 0): 0.30, (0, 1): 0.20, (1, 0): 0.15, (1, 1): 0.35}
Q = Query.interventional(outcome="Y", do={"X": 1}, value=1)
a = Assumptions.order("X", "Y") & Assumptions.mtr("Y", "X")
r = bound(V, P, Q, a)
print(r.lower, r.upper, r.tight)      # 0.55 0.85 True
\end{lstlisting}

\subsection{Objects}
\label{sec:design}

\texttt{Variables} declares names and finite domains; variables listed outside \texttt{observed} are latent endogenous variables that the query may still name.

The table is a dictionary of cell probabilities, or the plug-in estimate returned by \texttt{from\_dataframe} or \texttt{from\_counts}.

\texttt{Query} provides interventional probabilities, the ATE, the probabilities of causation (PNS, PN, PS), the counterfactual probability of a single event, and any linear combination of counterfactual events, possibly divided by an observational probability.

\texttt{bound} returns the interval, its width, the flags \texttt{tight} and \texttt{success}, the solver used, and notes on every reduction applied. Three companions share the same signature: \texttt{bound\_path} traces the interval along a sensitivity budget, \texttt{bound\_ci} adds a bootstrap confidence interval for the partially identified parameter \citep{imbens2004confidence}, and \texttt{classify\_program} reports the classification below without solving.

\subsection{Assumption factories}
\label{sec:factories}

Each factory returns a constraint set, and \texttt{\&} intersects them. Zero cells and linear inequalities keep the reduced program linear (\cref{prop:assumptions}); the last three rows of \cref{tab:catalog} change the solver.

\begin{table}[ht]
\centering
\caption{Assumption factories and the form of the constraint each one adds
to the program. Zero cells and linear inequalities keep the reduced program
linear (\cref{prop:assumptions}); the last three rows change the solver.}
\label{tab:catalog}
\small
\begin{tabular}{@{}>{\raggedright\arraybackslash}p{0.30\textwidth}>{\raggedright\arraybackslash}p{0.42\textwidth}l@{}}
\toprule
Factory & Assumption & Constraint \\
\midrule
\texttt{order}, \texttt{diagram}, \texttt{iv} & causal order or diagram
  \citep{manski1990nonparametric,balke1997bounds} & graph \\
\texttt{exclusion} & missing directed edge & zero cells \\
\texttt{mtr}, \texttt{no\_defiers} & monotone response
  \citep{manski1997monotone,imbens1994identification} & zero cells \\
\texttt{mts} & monotone selection \citep{manski2000monotone} & linear inequality \\
\texttt{tan\_msm}, \texttt{rosenbaum\_gamma} & odds-ratio budget, anchored
  \citep{tan2006distributional,rosenbaum2002observational} & linear inequality \\
\texttt{c\_dependence} & propensity budget \citep{masten2018identification} & linear inequality \\
\texttt{margin\_budget} & distance from no confounding & linear inequality \\
\texttt{f\_sensitivity} & $f$-divergence budget \citep{jin2023sensitivity} & convex \\
\texttt{support\_bound} & at most $k$ response types & mixed-integer \\
\texttt{independence} & independent latent blocks & product (outer bound) \\
\bottomrule
\end{tabular}
\end{table}

\subsection{What \texttt{bound} does}
\label{sec:solve}

With the default \texttt{method="auto"}, \texttt{bound} proceeds in three
steps.
\begin{enumerate}[leftmargin=1.4em,itemsep=0.2em]
\item \emph{Point identification.} On an explicit diagram with every
variable observed and no additional shape, sensitivity, or support restrictions, the identification algorithm of
\citet{shpitser2006identification} is run first. If it succeeds, the result
is the identifying functional evaluated on $P$, a degenerate interval.
\item \emph{Classification.} \texttt{classify\_program} restricts the graph
to the iterated closure of the query variables under directed ancestors and
$c$-components (\cref{prop:trim}); pins unconfounded variables
evaluated in a single world (\cref{lem:middle}); drops the $c$-components
the query does not read; and checks the three tests of \cref{thm:lp}.
It also reports the \emph{program degree}, the number of free latent
factors that remain.
\item \emph{Solve.} The class determines the solver and the meaning of the
returned interval (\cref{tab:result}).
\end{enumerate}

\begin{table}[ht]
\centering
\caption{Classification, solver, and what \texttt{bound} returns.
\texttt{tight=False} means that sharpness is not certified, not that the
interval is known to be loose.}
\label{tab:result}
\small
\begin{tabular}{@{}lll@{}}
\toprule
Classification & Solver & Returned interval \\
\midrule
point-identified & ID functional & point, \texttt{tight=True} \\
\cref{thm:lp} holds & closed form or LP (HiGHS) & sharp, \texttt{tight=True} \\
\cref{thm:lp} with a convex budget & convex program (CVXPY) & sharp, \texttt{tight=True} \\
\cref{thm:lp} with a support bound & mixed-integer LP (HiGHS) & sharp, \texttt{tight=True} \\
degree $2$ (bilinear) & nested search, else outer LP & \texttt{tight=False} unless certified \\
degree $\ge3$, \texttt{independence} & LP over merged latents & outer, \texttt{tight=False} \\
assumptions contradict $P$ & --- & \texttt{success=False} \\
\bottomrule
\end{tabular}
\end{table}

Inside the LP path the package merges response types that neither the observational map nor the query distinguishes, and switches to a dual or column-generation formulation when the type space is large. These steps reduce the size of the program without changing its optimum. Closed forms are used for Manski, Balke--Pearl, and a few anchored sensitivity models.

\paragraph{Scope.} The package targets discrete, finite systems. The table $P$ is treated as known; finite-sample uncertainty enters only through \texttt{bound\_ci}. It returns numbers, not symbolic bounds (\cref{sec:related}).

\clearpage

\section{Worked code for the literature catalogue}
\label{app:code}

This appendix gives runnable code for the assumption catalogue, the designs
in \cref{sec:illustrations}, the rows of \cref{tab:audit}, and further
instances from the literature: more treatment and outcome levels, two
treatments under one latent, and order-only chains. Each block is self-contained. Observational
tables come from \pid\texttt{.problems} or are written out; the graph,
query, and assumptions are explicit so that each block can serve as a
template for new data. The comment under each block is the printed output
(package version $0.1.0\mathrm{a}5$). Except for the last block, every
program below is certified LP or point by \cref{thm:lp}, and every interval
is \texttt{tight=True}.

\subsection{The assumption catalogue on the job-training table}
\label{app:code-zoo}

One table, one query, and every factory of \cref{tab:catalog} composed with
the causal order on~\eqref{eq:re1}. The order-only row is the natural bound
of \citet{manski1990nonparametric}; the other rows are the assumptions named
in that table.
\begin{lstlisting}
from partialid import Assumptions, Query, Variables, bound

V = Variables.from_binary("X", "Y")
P = {(0, 0): 0.30, (0, 1): 0.20, (1, 0): 0.15, (1, 1): 0.35}
Q = Query.interventional(outcome="Y", do={"X": 1}, value=1)
base = Assumptions.order("X", "Y")
catalogue = {
    "order (Manski)": base,
    "MTR": base & Assumptions.mtr("Y", "X"),
    "MTS": base & Assumptions.mts("X", "Y"),
    "MTR & MTS": base & Assumptions.mtr("Y", "X") & Assumptions.mts("X", "Y"),
    "Tan MSM, L=3": base & Assumptions.tan_msm("X", 3.0),
    "Rosenbaum, G=9": base & Assumptions.rosenbaum_gamma("X", 9.0),
    "c-dependence, c=0.1": base & Assumptions.c_dependence("X", 0.1),
    "margin, kappa=0.05": base & Assumptions.margin_budget(0.05),
    "f-sens. KL, rho=0.1": base & Assumptions.f_sensitivity("X", 0.1, f="kl"),
    "support k=3": base & Assumptions.support_bound(3),
}
for name, a in catalogue.items():
    r = bound(V, P, Q, a)
    out = f"[{r.lower:.3f}, {r.upper:.3f}]" if r.success else "infeasible"
    print(f"{name:22s} {out:16s} {r.method}")
# order (Manski)         [0.350, 0.850]   closed/manski_binary_xy
# MTR                    [0.550, 0.850]   fiber
# MTS                    [0.350, 0.700]   lp
# MTR & MTS              [0.550, 0.700]   lp
# Tan MSM, L=3           [0.467, 0.800]   closed/tan_msm_re1_L3
# Rosenbaum, G=9         [0.467, 0.800]   closed/anchored_rosenbaum_re1_G9
# c-dependence, c=0.1    [0.625, 0.750]   msm_agg/msm_transport_lp
# margin, kappa=0.05     [0.650, 0.750]   lp
# f-sens. KL, rho=0.1    [0.480, 0.846]   convex
# support k=3            infeasible       mip
\end{lstlisting}

\subsection{Ternary exposure}
\label{app:code-ternary}

A confounded treatment with three levels. Under the causal order alone the
natural bounds of \citet{manski1990nonparametric}, \eqref{eq:manski}, apply
for any finite $X$, with width $1-P(X{=}x)$.
\begin{lstlisting}
from partialid import Assumptions, Query, bound
from partialid.problems import ternary_exposure

p = ternary_exposure()            # X in {0,1,2}, Y in {0,1}
Q = Query.interventional(outcome="Y", do={"X": 1}, value=1)
a = Assumptions.order("X", "Y")   # natural bounds
r = bound(p.variables, p.data, Q, a)
print(f"[{r.lower:.3f}, {r.upper:.3f}] width={r.width:.3f} tight={r.tight} method={r.method}")
print(f"1 - P(X=1) = {p.expected_width:.3f}")
# [0.158, 0.730] width=0.573 tight=True method=fiber
# 1 - P(X=1) = 0.573
\end{lstlisting}

\subsection{Higher cardinality: four treatment levels}
\label{app:code-card42}

The same natural bound \citep{manski1990nonparametric}, now with $X\in\{0,1,2,3\}$.
\begin{lstlisting}
from partialid import Assumptions, Query, bound
from partialid.problems import ternary_cardinalities

p = ternary_cardinalities(card_x=4, card_y=2)   # X in {0,..,3}, Y in {0,1}
Q = Query.interventional(outcome="Y", do={"X": 1}, value=1)
r = bound(p.variables, p.data, Q, Assumptions.order("X", "Y"))
print(f"[{r.lower:.3f}, {r.upper:.3f}] width={r.width:.3f} tight={r.tight} method={r.method}")
print(f"1 - P(X=1) = {p.expected_width:.3f}")
# [0.023, 0.882] width=0.859 tight=True method=fiber
# 1 - P(X=1) = 0.859
\end{lstlisting}

\subsection{Higher cardinality: ternary treatment and outcome}
\label{app:code-card33}

The query targets one level of a ternary outcome. The same natural bound
of \citet{manski1990nonparametric} holds.
\begin{lstlisting}
from partialid import Assumptions, Query, bound
from partialid.problems import ternary_cardinalities

p = ternary_cardinalities(card_x=3, card_y=3)   # X, Y in {0,1,2}
Q = Query.interventional(outcome="Y", do={"X": 1}, value=1)   # P(Y_{x=1} = 1)
r = bound(p.variables, p.data, Q, Assumptions.order("X", "Y"))
print(f"[{r.lower:.3f}, {r.upper:.3f}] width={r.width:.3f} tight={r.tight} method={r.method}")
print(f"1 - P(X=1) = {p.expected_width:.3f}")
# [0.187, 0.861] width=0.674 tight=True method=fiber
# 1 - P(X=1) = 0.674
\end{lstlisting}

\subsection{Binary instrument}
\label{app:code-iv}

The encouragement design of \cref{sec:ill-designs,ex:iv-lp}, the binary
instrument of \citet{balke1997bounds}. The classifier reports
degree one after the directed cut $I=\{Z\}$, $R=\{X,Y\}$. Dropping the
instrument assumptions (order only) widens the interval.
\begin{lstlisting}
from partialid import Assumptions, Query, bound, classify_program
from partialid.problems import binary_iv

p = binary_iv(seed=2)
Q = Query.interventional(outcome="Y", do={"X": 1}, value=1)
a = Assumptions.iv("Z", "X", "Y")          # Z -> X -> Y, X <-> Y
r = bound(p.variables, p.data, Q, a)
print(f"[{r.lower:.3f}, {r.upper:.3f}] width={r.width:.3f} tight={r.tight} method={r.method}")
d = classify_program(a.diagram(), query=Q)
print(d.kind.value, d.degree, d.wl_wr_cut)

# the same table under order only (one latent on Z, X, Y)
r0 = bound(p.variables, p.data, Q, Assumptions.order("Z", "X", "Y"))
print(f"order only: [{r0.lower:.3f}, {r0.upper:.3f}]")
# [0.283, 0.615] width=0.333 tight=True method=closed/balke_pearl_iv
# linear 1 (('Z',), ('X', 'Y'))
# order only: [0.154, 0.705]
\end{lstlisting}

\subsection{Two instruments}
\label{app:code-two-iv}

Two independent instruments violate Condition~3 of
\citet[Proposition~2]{symbolicbounds2023general} but still admit the directed
cut $I=\{Z_1,Z_2\}$ (\cref{sec:sachs}).
\begin{lstlisting}
from partialid import Assumptions, Query, bound, classify_program
from partialid.problems import two_instruments

p = two_instruments()
Q = Query.interventional(outcome="Y", do={"X": 1}, value=1)
a = Assumptions.diagram(
    ["Z1", "Z2", "X", "Y"],
    directed=[("Z1", "X"), ("Z2", "X"), ("X", "Y")],
    bidirected=[("X", "Y")],
)
r = bound(p.variables, p.data, Q, a)
print(f"[{r.lower:.3f}, {r.upper:.3f}] width={r.width:.3f} tight={r.tight} method={r.method}")
d = classify_program(a.diagram(), query=Q)
print(d.kind.value, d.degree, d.wl_wr_cut)
# [0.397, 0.789] width=0.392 tight=True method=lp
# linear 1 (('Z1', 'Z2'), ('X', 'Y'))
\end{lstlisting}

\subsection{Two treatments under one latent}
\label{app:code-multi}

A multi-cause design of the kind treated by \citet{shridharan2023scalable}: two
treatments $T_0\to T_1$, both causing $Y$, with one latent shared by all three.
One hyperedge covers the ancestral set, so covering holds with a single hyperedge (\cref{thm:lp}).
\begin{lstlisting}
from partialid import Assumptions, Query, bound
from partialid.problems import shridharan_multi_treatment

p = shridharan_multi_treatment(n_treatments=2, card=2)   # T0, T1, Y
Q = Query.interventional(outcome="Y", do={"T0": 1}, value=1)
a = Assumptions.diagram(
    ["T0", "T1", "Y"],
    directed=[("T0", "Y"), ("T1", "Y"), ("T0", "T1")],
    bidirected=[("T0", "T1", "Y")],         # one latent on all three
)
r = bound(p.variables, p.data, Q, a)
print(f"[{r.lower:.3f}, {r.upper:.3f}] width={r.width:.3f} tight={r.tight} method={r.method}")
# [0.367, 0.794] width=0.428 tight=True method=fiber
\end{lstlisting}

\subsection{Order-only chains}
\label{app:code-chains}

Under a causal order $X\prec Z_0\prec\cdots\prec Y$ with maximal confounding,
the mediators are inert for $P(Y_x)$. The bound is the natural bound of
\citet{manski1990nonparametric} on the $(X,Y)$-margin, and the package prunes
the mediators before solving.
\begin{lstlisting}
from partialid import Assumptions, Query, bound
from partialid.problems import chain_order_only

p = chain_order_only(3, card=2)          # X, Z0, Y binary
Q = Query.interventional(outcome="Y", do={"X": 1}, value=1)
r = bound(p.variables, p.data, Q, Assumptions.order("X", "Z0", "Y"))
print(f"[{r.lower:.3f}, {r.upper:.3f}] width={r.width:.3f} tight={r.tight} method={r.method}")
print(f"1 - P(X=1) = {p.expected_width:.3f}")
# [0.445, 0.820] width=0.374 tight=True method=fiber
# 1 - P(X=1) = 0.374
\end{lstlisting}
A longer binary chain and a ternary chain change only the constructor and the
order:
\begin{lstlisting}
p = chain_order_only(4, card=2)          # X, Z0, Z1, Y binary
r = bound(p.variables, p.data, Q, Assumptions.order("X", "Z0", "Z1", "Y"))
# [0.241, 0.761] width=0.520 tight=True method=fiber   (1 - P(X=1) = 0.520)

p = chain_order_only(3, card=3)          # X, Z0, Y in {0,1,2}
r = bound(p.variables, p.data, Q, Assumptions.order("X", "Z0", "Y"))
# [0.095, 0.722] width=0.627 tight=True method=fiber   (1 - P(X=1) = 0.627)
\end{lstlisting}

\subsection{Front door}
\label{app:code-fd}

The front-door criterion of \citet{pearl1995diagrams}: $X$ affects $Y$ only
through $Z$, and the latent is shared by $X$ and $Y$. The functional is a point.
\begin{lstlisting}
from partialid import Assumptions, Query, Variables, bound

V = Variables.from_binary("X", "Z", "Y")
pz1, py1 = {0: 0.4, 1: 0.7}, {0: 0.3, 1: 0.8}      # P(Z=1|x), P(Y=1|z)
P = {(x, z, y): 0.5 * (pz1[x] if z else 1 - pz1[x]) * (py1[z] if y else 1 - py1[z])
     for x in (0, 1) for z in (0, 1) for y in (0, 1)}
Q = Query.interventional(outcome="Y", do={"X": 1}, value=1)
fd = Assumptions.diagram(["X", "Z", "Y"], directed=[("X", "Z"), ("Z", "Y")],
                         bidirected=[("X", "Y")])
r = bound(V, P, Q, fd)
print(f"[{r.lower:.3f}, {r.upper:.3f}] tight={r.tight} method={r.method}")
r0 = bound(V, P, Q, Assumptions.order("X", "Z", "Y"))
print(f"order only: [{r0.lower:.3f}, {r0.upper:.3f}]")
# [0.650, 0.650] tight=True method=id
# order only: [0.325, 0.825]
\end{lstlisting}

\subsection{Measurement error}
\label{app:code-me}

\begin{lstlisting}
from partialid import Assumptions, Query, Variables, bound

V = Variables.from_binary("X", "Y", "Y2", observed=["X", "Y2"])   # Y is latent
P = {(0, 0): 0.30, (0, 1): 0.20, (1, 0): 0.15, (1, 1): 0.35}      # table of (X, Y2)
Q = Query.ate(outcome="Y", treatment="X", y=1, t1=1, t0=0)
a = (Assumptions.diagram(["X", "Y", "Y2"], directed=[("X", "Y"), ("Y", "Y2")],
                         bidirected=[("Y", "Y2")])
     & Assumptions.mtr("Y2", "Y"))
r = bound(V, P, Q, a)
print(f"[{r.lower:.3f}, {r.upper:.3f}] tight={r.tight} method={r.method}")
# [-0.400, 1.000] tight=True method=lp
\end{lstlisting}

\subsection{The multi-site panel}
\label{app:code-panel}

\begin{lstlisting}
import pandas as pd
from partialid import Assumptions, Query, bound, bound_ci, from_dataframe

cols = ["site", "age_bin", "educ", "prior_emp", "urban", "spouse", "treated", "employed"]
df = pd.read_csv("notebooks/data/job_training_panel.csv")        # n = 5000
order = Assumptions.order(*cols)
Q = Query.interventional(outcome="employed", do={"treated": 1}, value=1)

d = from_dataframe(df, columns=cols)
r = bound(d.variables, d.P, Q, order)
print(f"pooled:   [{r.lower:.3f}, {r.upper:.3f}]")

Qu = Query.interventional(outcome="employed", do={"treated": 1}, value=1,
                          condition={"urban": 1})
r = bound(d.variables, d.P, Qu, order)
print(f"urban=1:  [{r.lower:.3f}, {r.upper:.3f}]")

s0 = df[df["site"] == 0]                                          # n = 1640
d0 = from_dataframe(s0, columns=cols)
a = order & Assumptions.mtr("employed", "treated")
r = bound(d0.variables, d0.P, Q, a)
ci = bound_ci(d0.variables, d0.P, Q, a, n=len(s0), n_boot=200, seed=0)
print(f"site 0, MTR: [{r.lower:.3f}, {r.upper:.3f}]  95% CI [{ci.lower:.3f}, {ci.upper:.3f}]")
# pooled:   [0.369, 0.862]
# urban=1:  [0.403, 0.860]
# site 0, MTR: [0.641, 0.866]  95% CI [0.622, 0.881]
\end{lstlisting}

\subsection{The directed chain under four latent structures}
\label{app:code-chain}

\begin{table}[ht]
\centering
\caption{Directed chain $X\to W\to Y$ on~\eqref{eq:re3}. The classification
is that of \texttt{classify\_program}, with the program degree in
parentheses.}
\label{tab:audit}
\small
\begin{tabular}{@{}lllll@{}}
\toprule
Latents & Query & Classification & Interval & \texttt{tight} \\
\midrule
$\{X,W\}$ & $P(Y_1{=}1)$ & LP (1) & $[0.470,0.570]$ & yes \\
$\{X,W\}$ & PNS & bilinear (2) & $[0.000,0.520]$ & no \\
$\{X,W\},\{W,Y\}$ & $P(Y_1{=}1)$ & LP (2) & $[0.270,0.770]$ & yes \\
$\{X,W\},\{W,Y\}$ & PNS & LP (2) & $[0.000,0.520]$ & yes \\
$\{X,W,Y\}$ & $P(Y_1{=}1)$ & LP (1) & $[0.270,0.770]$ & yes \\
three pairwise edges & $P(Y_1{=}1)$ & polynomial (3) & $[0.270,0.770]$ & no \\
\bottomrule
\end{tabular}
\end{table}

The rows of \cref{tab:audit}. The mediated PNS and the triangle are not
certified, and the code prints \texttt{tight=False} for them.
\begin{lstlisting}
from partialid import Assumptions, Query, Variables, bound, classify_program

V = Variables.from_binary("X", "W", "Y")
Pxw = {(0, 0): 0.25, (0, 1): 0.25, (1, 0): 0.15, (1, 1): 0.35}
pyw = {0: 0.4, 1: 0.6}                                            # P(Y=1|w)
P = {(x, w, y): Pxw[x, w] * (pyw[w] if y else 1 - pyw[w])
     for x in (0, 1) for w in (0, 1) for y in (0, 1)}
queries = {"P(Y_1=1)": Query.interventional(outcome="Y", do={"X": 1}, value=1),
           "PNS": Query.pns(treatment="X", outcome="Y")}
latents = {"mediated": [("X", "W")],
           "pair chain": [("X", "W"), ("W", "Y")],
           "one oval": [("X", "W", "Y")],
           "triangle": [("X", "W"), ("W", "Y"), ("X", "Y")]}
for name, bi in latents.items():
    a = Assumptions.diagram(["X", "W", "Y"], directed=[("X", "W"), ("W", "Y")],
                            bidirected=bi)
    for qn, Q in queries.items():
        k = classify_program(a.diagram(), query=Q, variables=V)
        r = bound(V, P, Q, a)
        print(f"{name:10s} {qn:8s} {k.kind.value:15s} {k.degree} "
              f"[{r.lower:.3f}, {r.upper:.3f}] tight={r.tight}")
# mediated   P(Y_1=1) linear          1 [0.470, 0.570] tight=True
# mediated   PNS      bilinear/factor 2 [0.000, 0.520] tight=False
# pair chain P(Y_1=1) linear          2 [0.270, 0.770] tight=True
# pair chain PNS      linear          2 [0.000, 0.520] tight=True
# one oval   P(Y_1=1) linear          1 [0.270, 0.770] tight=True
# one oval   PNS      linear          1 [0.000, 0.520] tight=True
# triangle   P(Y_1=1) polynomial      3 [0.270, 0.770] tight=False
# triangle   PNS      polynomial      3 [0.000, 0.520] tight=False
\end{lstlisting}

\clearpage

\section{Related work and software}
\label{sec:related}

\paragraph{Programs for discrete partial identification.}
\citet{balke1994bounds,balke1997bounds} wrote the instrumental-variable
problem as an LP over response functions and solved it symbolically.
Canonical SCMs extend the response-function parameterization to any
diagram \citep{zhang2022partial}, at the price of a polynomial program in
general; \citet{zhang2022partial} approach it by Bayesian sampling, and
\citet{duarte2024automated} by polynomial programming that returns sharp
bounds or valid outer bounds with a certified gap.
\citet{finkelstein2020bounds} derive bounds from logical relations among
counterfactuals. These works either fix one tractable design or give a sufficient condition
for a linear program. \Cref{thm:lp} is necessary and sufficient for diagrams
in which every $c$-component has at most two hyperedges: it says when the
program is an LP and when it is not, and the package uses that answer to pick
the solver and to label the result.

The same canonical model is also used without forming the program.
\citet{bellot2026foundation} put a full-support prior on canonical SCMs with discrete observables and train a prior-data fitted network \citep{muller2022transformers} that maps an observational table to a distribution over a counterfactual query.
The support of that distribution approximates the identified set.
\Cref{thm:lp} instead reads off the diagram and the query when the set is the optimum of one linear program.
Partial identification is also posed across domains.
\citet{jalaldoust2024partial} bound a functional of a target law, such as the risk of a fixed classifier, from source-domain data and a selection diagram, by gradient optimization over a structural parameterization.
The constraints there are invariances of mechanisms across domains, rather than the single observational table of \cref{def:reduced}.

\paragraph{Assumptions and sensitivity models.}
Shape restrictions \citep{manski1990nonparametric,manski1997monotone,manski2000monotone}
and sensitivity models \citep{rosenbaum2002observational,tan2006distributional,masten2018identification,jin2023sensitivity}
usually come with their own derivation and estimator, often for one
graph; see \citet{tamer2010partial} and \citet{kline2023recent} for reviews.
In the program of \cref{def:optimal_bound}, the anchored versions of these
models are linear constraints on the same response-type law
(\cref{prop:assumptions}), which is what lets the package compose them with each other
and with a diagram. Unanchored odds-ratio models and information-theoretic
budgets \citep{jung2026information} are not linear in this sense and are
outside the LP path.

\begin{table}[t]
\centering
\caption{Software for bounds on causal effects with discrete data.}
\label{tab:software}
\small
\begin{tabular}{@{}*{4}{>{\raggedright\arraybackslash}p{0.225\textwidth}}@{}}
\toprule
Tool & Language & Scope & Output \\
\midrule
\texttt{causaloptim} \newline \citep{symbolicbounds2023general} & R
  & graphs in their class, linear constraints & symbolic sharp bounds \\
Autobounds \newline \citep{duarte2024automated} & Python
  & any discrete graph, polynomial constraints & numeric bounds, optimality gap \\
CausalQueries \newline \citep{humphreys2023integrated} & R
  & discrete graphs, type restrictions & posterior over types \\
\texttt{bpbounds} \newline \citep{palmer2011nonparametric} & R, Stata
  & instrumental variables & closed-form Balke--Pearl bounds \\
\texttt{sensemakr} \newline \citep{cinelli2020making} & R, Stata, Python
  & linear regression & omitted-variable sensitivity \\
\pid{} & Python
  & discrete graphs or orders, named assumptions & numeric bounds, sharpness flag \\
\bottomrule
\end{tabular}
\end{table}

\paragraph{Software.}
\Cref{tab:software} separates tools that bound a counterfactual query in a
discrete structural model from tools that answer a different question.
\texttt{bpbounds} returns the closed-form Balke--Pearl interval for a binary
instrument. CausalQueries returns a posterior over response types.
\texttt{sensemakr} reports how a coefficient in a linear regression moves
when an omitted variable of a stated strength is introduced. The other three
packages take a discrete diagram and a causal query and return bounds.

\texttt{causaloptim} enumerates the vertices of one linear program and
returns a symbolic expression in the observational probabilities. The
expression applies to every table, for graphs that meet their
Proposition~2 and for constraints that stay linear in one joint type law.
\Cref{thm:lp} contains that class (\cref{sec:sachs}). Inside diagrams with
at most two hyperedges in each component, \cref{thm:lp} is an equivalence
and their Proposition~2 is a sufficient condition. \Cref{app:sachs} lists the
graphs on which \cref{thm:lp} is a linear program and their Proposition~2 is not.

Autobounds accepts an arbitrary discrete diagram and encodes the canonical
model as a polynomial program. \Cref{thm:lp} decides, before a solver is
called, when that program is linear. \Cref{app:autobounds} runs both packages
on the examples of \cref{sec:illustrations}.

\subsection{Comparison with Sachs et al.}
\label{app:sachs}

\citet[Proposition~2]{symbolicbounds2023general} give a sufficient condition
for a tight linear program in one joint type law, and \texttt{causaloptim}
returns a symbolic bound for the graphs that meet it. \Cref{sec:sachs}
records the four conditions and shows that every graph meeting them meets
\cref{thm:lp}. Their condition is sufficient. \Cref{thm:lp} is an equivalence
for diagrams whose components have at most two hyperedges, and the inclusion
of their class is one way. The rows below are the other direction: \cref{thm:lp} returns a linear program, and no
bipartition meets their conditions. The comparison is the criterion, not the
running time. The classical instrument is the reference case, in both
classes.

\begin{table}[ht]
\centering
\caption{Graphs on which \cref{thm:lp} is a linear program.
``In both'' means Proposition~2 of \citet{symbolicbounds2023general} applies
as well. The other rows have no bipartition meeting their conditions
(1)--(4). Intervals are the sharp bounds under \cref{thm:lp}.}
\label{tab:sachs}
\small
\begin{tabular}{@{}p{0.28\textwidth}lp{0.32\textwidth}l@{}}
\toprule
Graph & Prop.~2 & Why it fails & Interval \\
\midrule
Instrument, $X\leftrightarrow Y$ & in both & --- & $[0.283,0.615]$ \\
Mediated $X\leftrightarrow W\to Y$ & no & $Y$ is pinned, not confounded with $\{X,W\}$ & $[0.470,0.570]$ \\
Two instruments $Z_1,Z_2$ & no & $Z_1$ and $Z_2$ share no latent & $[0.397,0.789]$ \\
Instrument with $Z\leftrightarrow X$ & no & the latent on $\{Z,X\}$ crosses every cut & $[0.283,0.615]$ \\
Pair chain, PNS & no & two latents in one $c$-component & $[0.000,0.600]$ \\
\bottomrule
\end{tabular}
\end{table}

\paragraph{Mediated chain.}
The graph is $X\to W\to Y$ with one latent on $\{X,W\}$ and $U\perp U_Y$,
the query is $P(Y_1{=}1)$, and the table is~\eqref{eq:re3}. The event uses
$Y$ in one world, so \cref{lem:middle} pins $Y$ and \cref{thm:lp} leaves an
LP in $P(U)$ (\cref{ex:mediated}). The three bipartitions of $\{X,W,Y\}$ each
fail their conditions: $Y$ does not share the latent of $\{X,W\}$, the latent
on $\{X,W\}$ meets both sides of the cut after $X$, and $P(Y_x)$ is not a
linear functional of the types of $Y$ alone. Their algorithm therefore does
not apply. Forcing it by taking a free joint over $(R_X,R_W,R_Y)$ drops
$U\perp U_Y$ and widens the interval from $[0.470,0.570]$ to
$[0.270,0.770]$.

\paragraph{Two instruments.}
$Z_1,Z_2\to X\to Y$ with $X\leftrightarrow Y$ and $Z_1\perp Z_2$. The only
cut with all arrows pointing the same way puts both instruments on the left
and $\{X,Y\}$ on the right. Their condition~(3) asks the left side to share
one latent, and $Z_1$ and $Z_2$ do not. Both instruments are $c$-components
the query does not read, so \cref{thm:lp} still reduces the problem to the
Balke--Pearl LP in the response types of $\{X,Y\}$. On the packaged table the
interval is $[0.397,0.789]$.

\paragraph{A latent between the instrument and the treatment.}
Add the hyperedge $\{Z,X\}$ to the classical instrument, and keep
$\{X,Y\}$. The query $P(Y_1{=}1)$ does not change. Every bipartition either
lets the latent on $\{Z,X\}$ or the latent on $\{X,Y\}$ meet both sides, or
asks $\{Z,X,Y\}$ to share one latent. Proposition~2 does not apply.
\Cref{prop:lift} does: $Z$ is a root, the independence of the two latents is
linear, and the sharp interval equals the Balke--Pearl interval of the
classical instrument. On the seed-$2$ table both graphs give
$[0.283,0.615]$. Sharing a latent with the treatment does not change the
bound, and it takes the graph out of their class.

\paragraph{Pair chain.}
$X\to W\to Y$ with hyperedges $\{X,W\}$ and $\{W,Y\}$ is one $c$-component
with two latents (\cref{ex:pair}). The same case analysis leaves no
bipartition: a single latent does not cover all three nodes, and each
pairwise latent crosses the cut that would separate its endpoints.
$X$ is a root, so \cref{lem:two} makes the sharp program an LP for every
linear query, including PNS. On the table~\eqref{eq:re2} that LP gives
$[0.000,0.600]$. One latent on $\{X,W,Y\}$, which does meet their
conditions, gives the wider interval $[0.000,0.650]$. For $P(Y_1{=}1)$ on
this table the two models agree, at $[0.350,0.850]$; the probabilities of
causation are where the second latent binds.

\subsection{Comparison with Autobounds}
\label{app:autobounds}

Autobounds \citep{duarte2024automated} encodes the canonical model as a
polynomial program and optimizes it by spatial branch-and-bound. \Cref{thm:lp}
decides, before any solver is called, when that program is a
point or a linear program. This subsection runs both packages on the examples of
\cref{sec:illustrations} that are small enough for the general solver to
finish in seconds. The intervals agree. The difference is the time, and it
tracks the classification.

\begin{table}[ht]
\centering
\caption{One call of each package on the same table, graph, and query.
Times are wall-clock on a laptop. The Autobounds call for the mediated graph
was stopped at a limit of $25\,\mathrm{s}$; its primal and dual endpoints
agree, so the returned interval is certified. On the front door the two
Autobounds endpoints differ by about $0.002$ and both round to $0.65$.}
\label{tab:autobounds}
\small
\begin{tabular}{@{}lllr@{}}
\toprule
Example & Interval & \pid{} & Autobounds \\
\midrule
Bow, $P(Y_1{=}1)$ on~\eqref{eq:re1} & $[0.350,0.850]$ & $<1\,\mathrm{ms}$ & $0.6\,\mathrm{s}$ \\
Bow, monotone response & $[0.550,0.850]$ & $1\,\mathrm{ms}$ & $0.5\,\mathrm{s}$ \\
Markovian bow, $P(Y_1{=}1)$ & $0.700$ & $1\,\mathrm{ms}$ & $0.5\,\mathrm{s}$ \\
Instrument, seed $2$ & $[0.283,0.615]$ & $14\,\mathrm{ms}$ & $0.5\,\mathrm{s}$ \\
Front door & $0.650$ & $1\,\mathrm{ms}$ & $5.5\,\mathrm{s}$ \\
Mediated chain on~\eqref{eq:re3} & $[0.470,0.570]$ & $1\,\mathrm{ms}$ & $25\,\mathrm{s}$ \\
\bottomrule
\end{tabular}
\end{table}

On the bow, with or without monotone response, and on the instrument,
\cref{thm:lp} leaves a single simplex: one $c$-component of $\lambda=1$, or,
for the instrument, the unread component $\{Z\}$ dropped by the directed cut.
The sharp interval is then a linear program, and on the binary bow and the
binary instrument it is a closed form. Autobounds computes the same interval
by branch-and-bound over the canonical parameterization. The answers match to
three decimals, including the ATE rows omitted from \cref{tab:autobounds}
($[-0.350,0.650]$ under the order alone and $[0.000,0.650]$ under monotone
response), and the linear solver finishes in milliseconds rather than about
half a second.

The Markovian bow and the front door are the case $K=0$ of \cref{thm:lp}.
The identification algorithm of \citet{shpitser2006identification} returns a
functional of $P$, and \pid{} evaluates it. Autobounds does not
take that branch: it builds and searches a program whose optimum happens to
be a point. That search is cheap for the two-node bow and takes
$5.5\,\mathrm{s}$ for the front door.

The mediated chain is the instance where the theorem changes the program
rather than only the solver. The graph is $X\to W\to Y$ with one latent on
$\{X,W\}$, the table is~\eqref{eq:re3}, and the query is $P(Y_1{=}1)$. In the
latent laws the outcome contributes a second factor, $P(U_Y)$. The event uses
$Y$ in one world, so \cref{lem:middle} pins $Y$ to the kernel $P(Y\mid W)$
and \cref{thm:lp} leaves an LP in $P(U)$ (\cref{ex:mediated}).
\pid{} returns $[0.470,0.570]$ in about $1\,\mathrm{ms}$.
Autobounds optimizes the unreduced program and returns the same interval at
the $25\,\mathrm{s}$ limit, with the primal--dual gap closed. This is the
same graph that lies outside the symbolic class of
\citet{symbolicbounds2023general} (\cref{sec:sachs}).

The same comparison shows a difference in how assumptions are stated. Monotone
response is one named constraint, intersected with the causal order:
\begin{lstlisting}
a = Assumptions.order("X", "Y") & Assumptions.mtr("Y", "X")
bound(V, P, Query.interventional(outcome="Y", do={"X": 1}, value=1), a)
\end{lstlisting}
The Autobounds call for the same row builds the confounded bow by hand and
sets one response event to probability zero:
\begin{lstlisting}
dag = DAG("X -> Y, U -> X, U -> Y", unob="U")
pro = causalProblem(dag)
pro.add_assumption(pro.p("Y(X=1)=0&Y(X=0)=1"), "==", 0)
pro.set_estimand(pro.p("Y(X=1)=1"))
pro.load_data(summary=table)
pro.solve()
\end{lstlisting}
Further assumptions in \pid{} are further names on the same
object. \Cref{tab:catalog} is the list: an instrument, an anchored sensitivity
budget, monotone selection, a support bound. The operator \texttt{\&}
intersects them, and \cref{thm:lp} is re-checked on the intersection, so a
constraint that preserves linearity stays on the linear-program path. In
Autobounds each extra assumption is a polynomial constraint the user writes,
and the solver does not change when that constraint happens to be linear.

The comparison is confined to instances \cref{thm:lp} certifies. On a graph
the criterion rejects, such as the latent triangle in \cref{tab:audit},
\pid{} returns an outer interval with \texttt{tight=False}.
Autobounds' branch-and-bound is the general method there, and we did not run
it.

\clearpage

\stopcontents[appendix]

\end{document}